\documentclass[10pt,twocolumn,letterpaper]{article}

\usepackage[pagenumbers]{cvpr} 

\definecolor{codegreen}{rgb}{0,0.6,0}

\newcommand{\spacebeforesection}{\vspace{-0.5em}}
\newcommand{\spaceaftersection}{\vspace{-0.5em}}

\definecolor{cvprblue}{rgb}{0.21,0.49,0.74}
\usepackage{booktabs}
\usepackage{colortbl}
\usepackage{svg}
\usepackage{xcolor}
\usepackage[pagebackref,breaklinks,colorlinks,allcolors=cvprblue]{hyperref}
\usepackage{float}
\usepackage{pifont}
\usepackage{amssymb} 
\usepackage{adjustbox}
\usepackage{graphicx}
\usepackage{float}  
\usepackage{multirow}
\definecolor{chrome-yellow}{RGB}{249, 219, 120} 
\definecolor{navy}{rgb}{0.0, 0.0, 0.5}
\def\paperID{5790} 
\def\confName{CVPR}
\def\confYear{2025}

\usepackage{color,soul}
\newcommand{\modelname}{LLAVIDAL}
\newcommand{\datasetname}{ADL-X}
\usepackage{svg}
\definecolor{LightBlue}{rgb}{0.68, 0.85, 0.9}
\definecolor{LightGreen}{rgb}{0.34, 0.91, 0.48}

\usepackage{calc}
\title{
    \begin{minipage}{0.125\textwidth}
        \includegraphics[width=\textwidth]{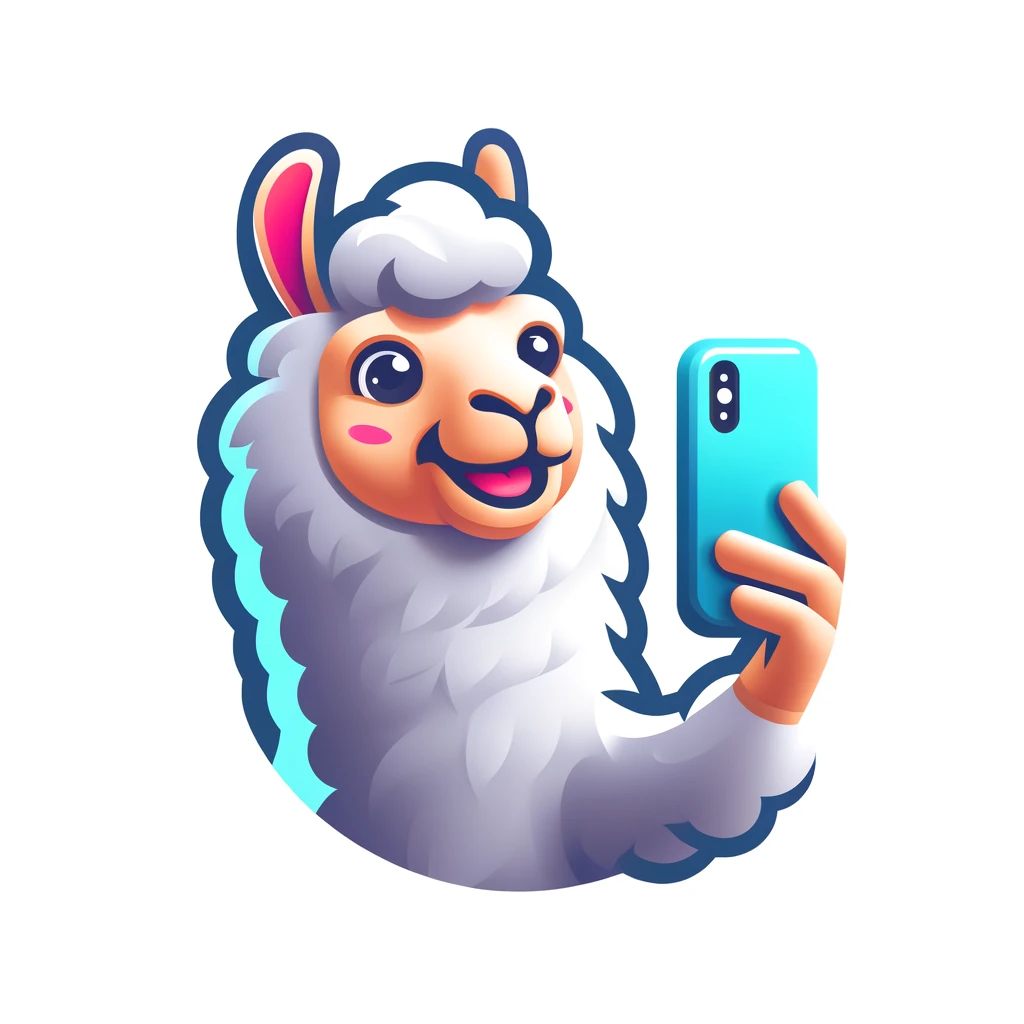}
    \end{minipage}
    \hspace{-0.15\textwidth}
    \begin{minipage}{0.85\textwidth}
        \centering
        \modelname~: A \underline{L}arge \underline{LA}nguage \underline{VI}sion Model \\ for \underline{D}aily \underline{A}ctivities of \underline{L}iving
    \end{minipage}
}

\author{
\textbf{Dominick Reilly$^{1}$} \hspace*{0.5em}
\textbf{Rajatsubhra Chakraborty$^{1}$} \hspace*{0.5em}
\textbf{Arkaprava Sinha$^{1}$} \hspace*{0.5em}
\textbf{Manish Kumar Govind$^{1}$} \hspace*{0.5em} \\
\textbf{Pu Wang$^{1}$} \hspace*{0.5em}
\textbf{Francois Bremond$^{3,4}$} \hspace*{0.5em}
\textbf{Le Xue$^{2}$} \hspace*{0.5em}
\textbf{Srijan Das$^{1}$} \hspace*{0.5em}
\vspace*{0.25em}
\\
$^{1}$ UNC Charlotte \hspace*{0.1em}
$^{2}$ Salesforce AI Research \hspace*{0.1em}
$^{3}$ Inria \hspace*{0.1em}
$^{4}$ Université Côte d'Azur
\vspace*{0.25em}
\\
{\tt\normalsize \{dreilly1,sdas24\}@charlotte.edu}
}

\begin{document}

\twocolumn[{
\renewcommand\twocolumn[1][]{#1}%
\maketitle
    \vspace{-10pt}
    \centering
    \scalebox{0.95}{
    \captionsetup{type=figure}
    \includegraphics[width=\textwidth]{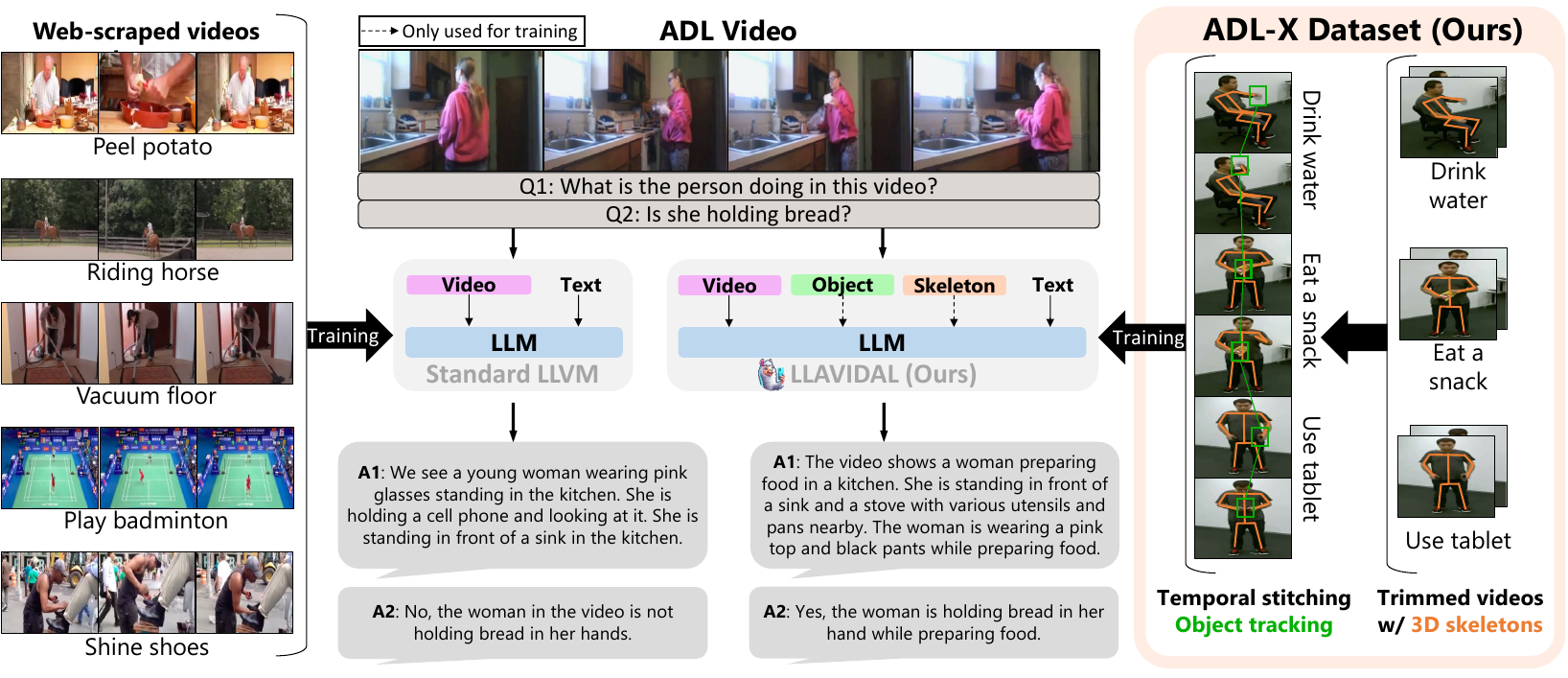}
}
\setcounter{figure}{0} \vspace{-0.1in}
    \captionof{figure}{Web-video trained Large Language Vision Models (LLVM) struggle to understand the fine-grained details and human-object interactions present in Activities of Daily Living (ADL). We propose \modelname, an LLVM trained with \textbf{three modalities} on our curated \textbf{ADL-X dataset}. ADL-X is derived from trimmed, multi-view ADL videos and is augmented with skeleton and object modalities.
    }
    \vspace{1em}
    \label{fig:teaser}
}]


\begin{abstract}
Current Large Language Vision Models (LLVMs) trained on web videos perform well in general video understanding but struggle with fine-grained details, complex human-object interactions (HOI), and view-invariant representation learning essential for Activities of Daily Living (ADL). This limitation stems from a lack of specialized ADL video instruction-tuning datasets and insufficient modality integration to capture discriminative action representations. To address this, we propose a semi-automated framework for curating ADL datasets, creating \textbf{\datasetname}, a multiview, multimodal RGBS instruction-tuning dataset. Additionally, we introduce \textbf{\modelname}, an LLVM integrating videos, 3D skeletons, and HOIs to model ADL's complex spatiotemporal relationships. For training \modelname~a simple joint alignment of all modalities yields suboptimal results; thus, we propose a Multimodal Progressive (\textbf{MMPro}) training strategy, incorporating modalities in stages following a curriculum. We also establish ADL MCQ and video description benchmarks to assess LLVM performance in ADL tasks. Trained on \datasetname, \modelname~achieves state-of-the-art performance across ADL benchmarks. Code and data will be made publicly available at \href{https://adl-x.github.io/}{https://adl-x.github.io/}.
\end{abstract}    
\vspace{-3em}
\section{Introduction}
\spaceaftersection
\label{sec:intro}

This paper aims explores the potential of Large Language-Vision Language Models (LLVMs) to understand Activities of Daily Living (ADL) which present various challenges including multiple exo-centric viewpoints, fine-grained activities with subtle motion, complex human-object interactions, and long-term temporal relationships.
Recently, ~\cite{videochat, videollava, videollama, videochatgpt, timechat, kevin2022egovlp,Yang_2023Vid2seq} have integrated videos into LLMs, leading to the development of video-based LLVMs capable of capturing spatio-temporal features. 
However, these models are predominantly trained on large-scale web videos~\cite{kinetics, something}, which mainly consists of sports clips~\cite{caba2015activitynet}, movie scenes~\cite{webvid_data, Tapaswi2015MovieQAUS}, and instructional videos~\cite{ego4d, howto100m, sener2022assembly101}.
These videos, typically filmed by professionals, follow strict temporal sequences in closely controlled background (e.g., \textit{Paragliding}). The evident temporal structure and scene semantics in such videos facilitate spatial understanding within LLVMs, as shown in Figure~\ref{fig:teaser}.
In contrast, ADL videos pose additional challenges, characterized by lack of a strict temporal structure where diverse actions may unfold concurrently within a single sequence~\cite{negin2019unsupervised}. For instance, \textit{a person cooking could intermittently engage in unrelated activities like making a phone call or drinking water, disrupting the linear progression of the composite action cooking}. Consequently, existing LLVMs trained on web videos struggle to capture such visually perplexing dynamics inherent in ADL scenarios. Moreover, unlike specialized video architectures designed for understanding ADL~\cite{glimpse, Baradel_BMVC, das2020vpn, vpn++, pivit, Gupta2015CrossMD, video_as_space_time}, these LLVMs lack explicit utilization of cues like 3D skeletons or human-object interactions (HOIs), which are crucial for understanding ADL. These cues facilitate the learning of view-invariant representations and capture fine-grained details essential for interpreting complex human activities.
Hence, the current limitations in understanding ADL stem from the lack of instruction tuning of LLVMs on real-world multiview ADL datasets captured in indoor settings and the simplistic design of LLVMs with holistic operations.

To address these challenges, we first propose a semi-automated framework for curating ADL video instruction tuning data. 
Unlike existing framework~\cite{videochatgpt, VideoLLM-online}, we introduce  novel strategies to guide AI annotators in focusing on  human-localized spatial regions, incorporating temporal unstructuredness in training distribution, and minimizing hallucinations through weak supervision to generate of high-quality video instruction pairs for training of LLVMs.
The result of this framework is \textbf{\datasetname}~dataset, comprising 100K untrimmed RGB video-instruction pairs, 3D skeletons (S), language descriptions. 
We then introduce the \textbf{L}arge \textbf{LA}nguage \textbf{VI}sion model for \textbf{D}aily \textbf{A}ctivities of \textbf{L}iving (\textbf{\modelname}), trained on \datasetname.  \modelname~integrates synchronized multimodal inputs from videos, including 3D skeletons and HOI (Human-Object Interaction) cues, into the LLM embedding space. Our study explores multiple strategies for multimodal integration in \modelname, addressing the challenge of jointly aligning all modalities with the LLM embeddings. To tackle this, we propose a Multimodal Progressive (MMPro) training strategy, enabling effective training of \modelname.
Furthermore, we introduce the \textbf{ADL Multiple Choice Question} (ADL MCQ) and \textbf{ADL video description} benchmarks, specifically designed to evaluate LLVM effectiveness in understanding ADL. Empirical results show that \modelname~outperforms other LLVMs, including those trained on datasets ten times larger, on these ADL benchmarks.
To summarize our contributions: 
\begin{itemize}
    \item We introduce \datasetname, the first multiview RGBS instruction tuning dataset for ADL tasks, curated via a novel semi-automated framework for training LLVMs. 
    \item We present \modelname, the first LLVM designed for ADL, integrating 3D skeletons and HOI cues into the LLM embedding space, and propose the \textit{MMPro} training strategy for synchronized multimodal training. \modelname~trained on \datasetname~outperforms all baseline LLVMs.
    \item We establish new benchmarks, ADL MCQ and video description tasks, to assess ADL understanding in LLVMs.
\end{itemize}

\spacebeforesection
\section{Related Works}
\spaceaftersection
Towards emulating human cognitive perception in digital intelligence, initial efforts focused on integrating vision and language modalities~\cite{vificlip, singh2022flava, XCLIP, FROSTER}.  Subsequently, the success of LLMs like GPT~\cite{gpt}, PALM~\cite{palm}, BLOOM~\cite{bloom} led to the introduction of multimodal conversational models\cite{minigpt, llama, llava, mplugowl,flamingo,pali,qwen} that combine image pixels and LLMs, we call LLVMs. In this section, we briefly discuss methodologies relevant to developing video conversational models. 


\textbf{Image captioners + LLM}.
With the emergence of foundation models~\cite{CLIP, zhai2023siglip, UniFormer, internvideo}, a natural extension is to use pre-trained VLM encoders to map visual inputs into language representations for LLM processing. For instance, CogVLM~\cite{cogvlm} uses this approach for image captioning, while Socratic Models~\cite{socratic} and VideoChat~\cite{videochat} adapt it for video captioning. Similarly, dialog-based models like VideoChatCaptioner~\cite{ChatCaptioner} summarize videos via interactions between ChatGPT~\cite{gpt} and a captioner such as BLIP2~\cite{blip2}, while ChatVideo~\cite{ChatVideo} uses task-specific foundation models with ChatGPT~\cite{gpt} to generate responses to user queries. Efficient video processing approaches, which encode video segments or frames via VLMs followed by single or multiple LLM calls with temporal hierarchy, have been extensively explored in~\cite{llovi, LongVLM, Kahatapitiya2024langrepo, kanchana_onepass, Park2024TooMF}. However, these models rely on static image-to-language mappings, lacking explicit modeling of temporal information.

\textbf{LLVMs}.
To align image pixels with the LLM embedding space, image-based methods~\cite{minigpt, mplugowl, instructblip, llava, qwen} employ a visual connector, often a simple linear projection layer. In contrast, models like Flamingo~\cite{flamingo} and BLIP2~\cite{blip2} use cross-attention mechanisms, such as a Query Transformer (Q-Former), to align visual features with textual prompts. For video alignment, LLVMs similarly use either linear projectors~\cite{videochatgpt, VideoLLM-online, jin2023_chatunivi} or Q-Formers~\cite{videochat, videollama, timechat} for feature encoding and alignment. However, these LLVMs generally capture a holistic view without focusing on fine-grained details or learning view-invariant representations, which are essential for understanding ADLs. In \modelname, additional video-derived modalities enable its visual connector to address these challenges.

\textbf{More Modalities + LLVMs}.
Methods like VideoLLama~\cite{videollama} and VideoLLaVA~\cite{videollava} incorporate additional modalities such as audio and images alongside videos. Similarly, X-InstructBLIP~\cite{panagopoulou2023xinstructblip}, aligns multiple modalities individually with language embedding space, as in LanguageBind~\cite{LanguageBind}. In contrast, we focus on synchronously aligning multiple modalities, requiring joint rather than separate alignment. This is achieved through our \textit{MMPro} training strategy, which uses progressive training~\cite{li2022autoprog, shen2023progressive} to sequentially integrate modalities into the LLM embedding space.

   

\begin{figure*}[t]  
    \centering
    \includegraphics[width=0.98\linewidth]{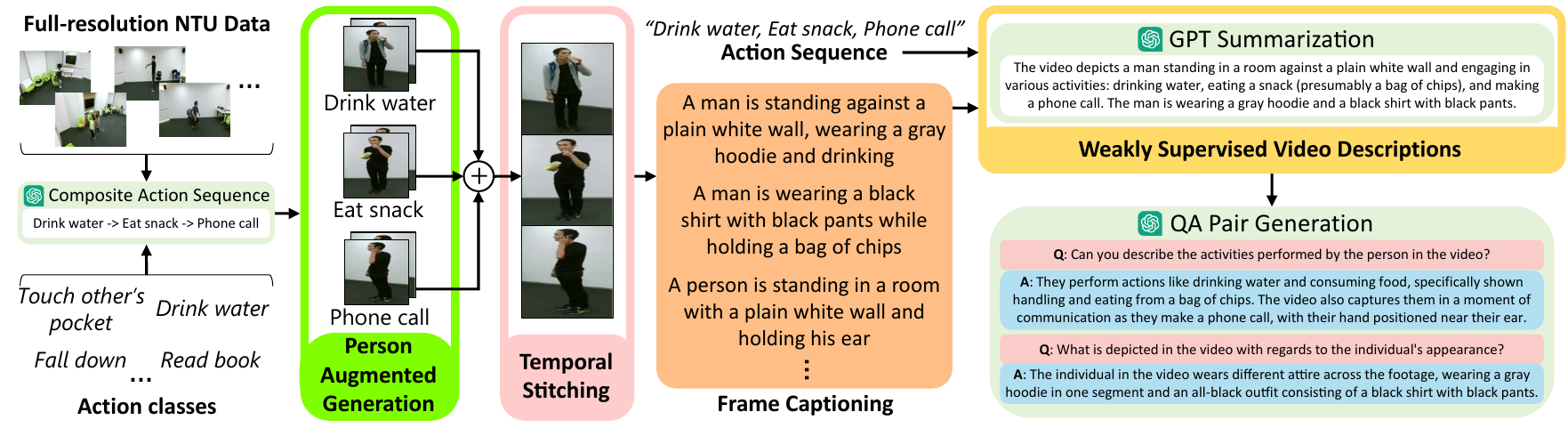}
    
    \caption{\textbf{ADL-X dataset curation pipeline.} The ADL-X dataset is derived from the NTU RGB+D 120 dataset through the use of three techniques: \colorbox{green}{Person Augmented Generation}, \colorbox{pink}{Temporal Stitching}, and \colorbox{chrome-yellow}{Weakly Supervised Video Descriptions}. The pipeline leverages CogVLM~\cite{cogvlm} for frame-level caption generation and GPT-3.5 Turbo~\cite{gpt} for summary synthesis and question-answer pair generation.}\vspace{-0.2in}
    \label{fig:data_curation}
\end{figure*}

\spacebeforesection
\section{Background: Training LLVM}
\spaceaftersection
In this section, we briefly present the general framework for training LLVMs. Given an input video $V_i \in \mathbb{R}^{T \times H \times W \times C}$, we encode each frame using the pretrained vision-language model (VLM) CLIP-L/14 \cite{CLIP} to obtain frame-level embeddings $f_i \in \mathbb{R}^{T \times h \times w \times D_{v}}$, where $D_{v}$ is the embedding dimension, and $h = H/p$, $w = W/p$ are dimensions adjusted by the patch size $p$. We extract temporal and spatial features by aggregating these embeddings along their respective dimensions \cite{videochatgpt}. The resulting video-level features, $\mathcal{X}^v_i \in \mathbb{R}^{F_{v} \times D_{v}}$, where $F_{v}$ denotes the number of spatio-temporal tokens, are obtained by concatenating the aggregated features.

These video features are projected into the LLM embedding space using a parameterized module $\mathcal{T}_v$, mapping input visual tokens to $Q_v \in \mathbb{R}^{F_{v} \times K}$. Similarly, the text query is tokenized to $Q_t \in \mathbb{R}^{F_t \times K}$, representing an instructional query from the training data. The inputs to the LLM are then combined as follows
$
[\text{USER:} \hspace{0.05in} \langle Q_t \rangle \hspace{0.05in} \langle Q_v \rangle \hspace{0.05in} \text{Assistant:}]
$.
Optimization of the LLVM is performed under a causal language modeling objective:
$\min_{\theta} L_{CE}(\textrm{LLM}(\mathcal{T}_v(\mathcal{X}^v_i)), y_i)$
where \(L_{CE}\) denotes the cross-entropy loss, \(\theta\) are the parameters of $\mathcal{T}_v$, \(y_i\) is the target sequence, and $\textrm{LLM}(\mathcal{T}_v(\mathcal{X}^v_i))$ is the LLM's prediction. 

\spacebeforesection
\section{ADL Video-instruction Pairs (\datasetname)}
\spaceaftersection
This section describes the weakly supervised semi-automated data curation framework employed for the creation of a novel dataset, \datasetname. This curation framework is carefully designed to enable the instruction tuning of LLVMs within the ADL domain.
\datasetname~comprises video recordings of ADLs from the NTU RGB+D 120 (NTU) dataset~\cite{ntu120}. This selection was motivated by the dataset's focus on ADL videos and its inherent diversity in terms of actions, subjects, and camera viewpoints.

To facilitate LLVM training, question-answer (QA) pairs pairs were systematically generated targeting various aspects of ADL.
The semi-automated data curation process for \datasetname~incorporates \textit{three} novel techniques as illustrated in Figure~\ref{fig:data_curation}: (i) Person Augmented Generation, which enhances video descriptions with spatially focused human related contents; (ii) Temporal Stitching, which connects short action clips for continuity; and (iii) Weakly Supervised Video Descriptions, which facilitates the generation of high-quality descriptions that guide the creation of video QA pairs. These QA pairs address diverse dimensions of ADLs, including human pose configurations, objects related to human actions, scene appearance, and the fine-grained actions performed. Subsequent sections detail the chronological steps undertaken to develop \datasetname.
\textbf{Person Augmented Generation (\colorbox{green}{PAG}).} ADL tasks necessitate focusing spatially on the individual's postures and their interactions with objects, distinct from the contextual background typical in web videos \cite{das2020vpn}. Specifically, we utilize skeleton data to crop bounding boxes around individuals. This person augmentation strategy minimizes background information in the video frames, effectively diminishes the generation of extraneous content by LLMs in video descriptions, focusing on human actions.

\textbf{Temporal Stitching (\colorbox{pink}{TS}).} 
Real-world ADL videos videos typically lack temporal structure, in contrast to instructional videos like \textit{cooking}, where actions are sequentially linked. To mimic the inherent randomness of ADLs, we develop a methodology to construct long, untrimmed video sequences by stitching together shorter clips. Initially, we generated 160 composite action sequences using GPT prompts to combine individual actions from the NTU dataset's 120 actions ($A_1$, $A_2$, ..., $A_{120}$). An example of such a sequence is \textit{drink water} $\rightarrow$ \textit{eat snack} $\rightarrow$ \textit{wipe face}.
The corresponding short clips from the NTU dataset are then stitched together, ensuring that all clips within a sequence maintain the same subject for consistency. To increase diversity and minimize bias towards specific subject-action pairings, we randomize both the sequence of actions and the subject assignments. This procedure yields \textbf{16,206 stitched videos}, with a maximum of 7 actions per video.

\textbf{Weakly Supervised (\colorbox{chrome-yellow}{WS}) Video Descriptions.} 
The generation of video descriptions, integral to automating QA pair creation, involves a two-step process: initially, image captions are generated for each frame in the video at a rate of $0.5$ fps using CogVLM~\cite{cogvlm}. 
Subsequently, these frame-level captions are synthesized into a cohesive video description via one LLM call. However, this step is prone to introducing hallucinations, potentially degrading the dataset's quality. To mitigate this, we employ weak supervision by incorporating the action sequence obtained from the short clips into the GPT-3.5 Turbo model, guiding it to generate a structured, coherent description limited to 300 words.
This method of weak supervision helps eliminate irrelevant noise from the caption generation process. For details on our prompting strategy, see Appendix~\ref{sec:prompts}.

\textbf{Generating QA Pairs.}
To generate domain-specific QA pairs for ADL, we utilize the dense video descriptions produced in the preceding step. An instruction template (detailed in Appendix~\ref{sec:prompts}), guides the GPT-3.5 model to formulate questions across various ADL-relevant categories. These categories include video summary, performed actions, spatial details, human-object interactions and other video-specific inquiries. Through this prompting approach, we curated a dataset comprising \textbf{100K video instruction pairs}, named \datasetname, derived from the stitched ADL videos.
Note that this data curation framework, applied to \datasetname, is adaptable and can be extended to other existing datasets for domain-specific training of LLVMs.

\spacebeforesection
\section{Multi-modalities of \datasetname} 
\spaceaftersection
\label{multimodal}
Following the literature for understanding ADL~\cite{Baradel_BMVC, pivit}, we recognize the importance of integrating additional modalities $\mathcal{M}_m$, such as 3D skeletons ($\mathcal{M}_s$) and human-object interactions ($\mathcal{M}_o$), to enhance the understanding of ADLs. These modalities are critical as ADLs predominantly involve movements of essential body parts or joints, facilitating the learning of view-invariant representations. Moreover, understanding not only the semantics of objects but also their trajectories — integral to the actions performed — is essential for developing fine-grained representations.
However, the question remains: how should these modalities be integrated within the LLM of the LLVM?
\begin{figure*}[t]
    \centering
    \includegraphics[width=0.96\linewidth]{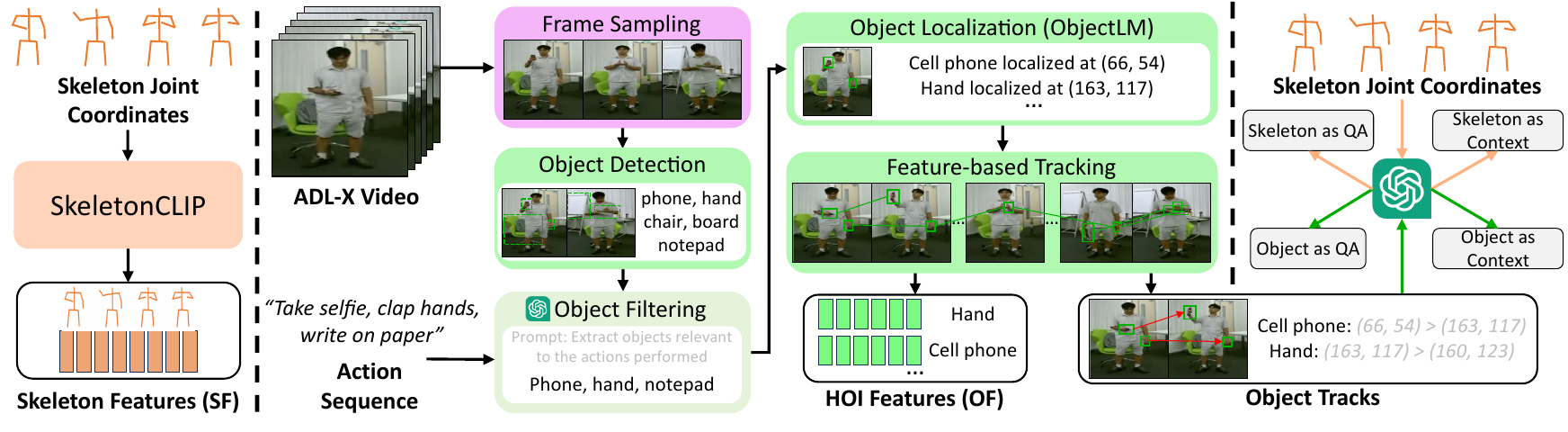}
    \vspace{-0.1in}
    \caption{\textbf{The multiple modalities of ADL-X.\quad Left:} Extraction of skeleton data features ($\mathcal{M}_s$) using SkeletonCLIP; \textbf{Middle:} Pipeline for extracting Human-Object Interaction (HOI) features through action-conditioned object detection, localization, and tracking; \textbf{Right:} Outline of our approach for obtaining $\mathcal{M}_m$ as QA and $\mathcal{M}_m$ as context.}\vspace{-0.2in}
    \label{fig:multimodal}
\end{figure*}

To address this, we investigate several strategies to incorporate skeletons and object tracks into the LLM input space. These include utilizing features extracted from dedicated modality-language encoders, augmenting QA pairs with modality-specific information, and contextualizing inputs to the LLM with modality-specific language as illustrated in Figure~\ref{fig:multimodal}.

\noindent \underline{\textbf{$\mathcal{M}_m$ as features.}}
To integrate new modalities into the LLVM input, we map modality-specific data from $\mathcal{M}_m \rightarrow \mathcal{X}_i^m$, which is combined with visual features $\mathcal{X}_i^v$. It is critical that the features $\mathcal{X}_i^m$, extracted from skeletons and HOI, are aligned to the language domain to facilitate their integration with the LLM embeddings, analogous to visual features derived from VLM~\cite{clip_representation}. We provide a detailed methodology for extracting these modality-specific features $\mathcal{X}_i^m$.

\textbf{Skeleton Features} (SF) - To extract features from the skeleton data $\mathcal{M}_s$ to be fed as input to the LLVM, we employ a skeleton-language model, specifically SkeletonCLIP~\cite{sinha2025skimodels}. SkeletonCLIP is a dual-encoder framework that combines a skeleton backbone~\cite{zhou2022hyperformer} and a frozen CLIP text encoder~\cite{CLIP}. Initially, the skeleton backbone is pretrained on trimmed NTU clips~\cite{NTU_RGB+D} for skeleton action classification. Subsequently, it is fine-tuned to enhance the alignment between skeleton features and language descriptions of actions using cross-entropy supervision. The resulting skeleton features are denoted as $\mathcal{X}^s_i \in \mathbb{R}^{F_s \times D_s}$, where $D_s$ indicates the dimension of skeleton features. These features are used as input tokens to LLVMs.

\textbf{HOI Features} (OF) - 
Extracting HOI features involves two steps: (i) \textit{Action-conditioned object detection} and (ii) \textit{Object Localization and Tracking}. Both steps leverage off-the-shelf models that are effective without the need for additional training. 
Given a stitched ADL video composed of a sequence of trimmed video segments (denoted as $\textrm{clip}_j$), the initial step involves extracting action-conditioned object categories from each clip. This is achieved by uniformly sampling 8 frames from each video and employing a pre-trained BLIP-2 model~\cite{blip2} to generate a list of distinct objects observed in the frames. Subsequently, we refine this list using the ground-truth action labels and GPT-3.5. Specifically, for each $\textrm{clip}_j$ in a stitched video, the corresponding action label and the list of detected objects are input into GPT-3.5, which is prompted to identify the object(s) most relevant to the given action. For example, if the objects \textit{plant, chair, bottle, table} are detected in a video labeled with the action \textit{Drinking}, GPT-3.5 is expected to filter out and select [\textit{bottle}] as the relevant object. Refer to Appendix~\ref{sec:prompts} for our detailed action-conditioned object detection prompting strategy.

In the second step, we perform the spatial localization and temporal evolution (i.e., tracking) of these identified objects within the stitched video. We employ a pre-trained open vocabulary object localization model, \textbf{ObjectLM} (OWLv2~\cite{OWLv2}), inputting the list of relevant objects along with the corresponding video. For each sampled frame, object bounding boxes are detected, and features for each object are extracted from the image regions within these boxes using ObjectLM. We denote the features for $n$ objects in frame $t$ as $\mathcal{X}_o^t \in \mathbb{R}^{n \times D_o}$, where $D_o$ represents the object feature dimension.
To track objects across frames, for each object in frame $t$, we compute the cosine similarity between its feature vector $\mathcal{X}_o^{t}$ and all feature vectors in frame $t+1$ corresponding to the same object category. This object in frame $t$ is then associated with the object in frame $t+1$ that exhibits the highest similarity score. This matching process is iterated for all objects across each frame, establishing a track for each relevant object throughout the sampled frames. Consequently, for $n$ relevant objects detected across 8 frames, the object features are structured using the following template:
$
[
\langle \mathcal{X}_o \rangle = \langle \mathcal{X}_o^1 \rangle \hspace{0.05in} \langle \mathcal{X}_o^2 \rangle \hspace{0.05in} ... \hspace{0.05in} \langle \mathcal{X}_o^n \rangle
]
$
where $\mathcal{X}_o^j \in \mathbb{R}^{8 \times D_{o}}$ represent the features of each tracked relevant object which are the HOI features in the video.

\noindent \underline{$\mathbf{\mathcal{M}_m}$ \textbf{as QA.}} 3D skeleton joint coordinates or relevant object trajectory coordinates are input alongside the associated action sequence into GPT-3.5 Turbo~\cite{gpt}, which generates a general description of the skeleton motion or human-object interactions (HOI) of an ADL-X video. This description is then re-fed into GPT-3.5 Turbo to generate two QA pairs that provide detailed explanations of the skeleton and object motions. These QA pairs are then added to the training set of text queries, $Q_t$, to tune the LLVM instruction.

\noindent \underline{\textbf{$\mathcal{M}_m$ as context.}} To integrate contextual information of human skeletons or human-object interactions, we append modality-specific information to the input text query $Q_t$ while training the LLVM.
For $\mathcal{M}_s$, we initially identify five peripheral joints -- the head, right hand, left hand, right knee, and left knee -- due to their significant contribution to motion. For $\mathcal{M}_o$, we utilize the trajectory coordinates of the relevant object(s) in the videos. Using GPT-3.5 Turbo, we generate descriptions of the motion for each of these joints or objects based on their trajectories throughout the video, specifically focusing on how the joint and object coordinates evolve. The generated descriptions, denoted as $Q^m_t | m=\{s, o\}$, are subsequently appended to the text query $Q_t$, incorporates these skeleton or human-object descriptions as additional contextual information. This enriched query $Q_t^{new} = [Q^m_t \hspace{0.05in} Q_t]$ is then employed for instruction tuning. 

\vspace{-0.8em}
\section{Multimodal Progressive (MMPro) Training}
\spaceaftersection

We dub our trained LLVM on on \datasetname~as \modelname. Integrating QA pairs and contextual information into $Q_t$ is achieved using conventional training methodologies for LLVMs. However, the joint integration of time-synchronized modalities ($\mathcal{X}_v$, $\mathcal{X}_s$, and $\mathcal{X}_o$) presents challenges, primarily due to conflicting gradients from the different modalities. To address this, we utilize modality-specific connectors that align each modality with the LLVM input space.
To mitigate the challenges of training with multiple modalities, we adopt a Multimodal Progressive (MMPro) training strategy for \modelname. This approach incrementally increases the training complexity by progressively adding modality-specific connectors, following a predefined growth schedule. These connectors project the modality-specific features into the LLVM embedding space, facilitating effective multimodal integration.

MMPro training is structured into $|\eta|$ equispaced stages, with $\frac{\#\text{Total iterations}}{|\eta|}$ iterations per stage. In the case of \modelname, where we integrate three modalities via connectors into the LLM embedding space, $\eta = 3$ stages. During stage 1, alignment of specific-modality with LLM embedding space is performed. Consequently, video, skeleton, and HOI features are independently projected into the LLM embedding space using linear projection layers $\mathcal{T}_m$ and their respective parameters $\theta_m$ for each cue $m=\{v,s,o\}$, resulting in LLM input token representations of the video, skeleton, and HOI cues, respectively:
\vspace{-0.6em}
\begin{equation}
     \hspace{-0.1in} Q_v = \mathcal{T}_v(\mathcal{X}_v; \theta_v); \hspace{0.05in} Q_s = \mathcal{T}_s(\mathcal{X}_s; \theta_s); \hspace{0.05in} Q_o = \mathcal{T}_o(\mathcal{X}_o; \theta_o)\vspace{-0.1in}
\end{equation}
where \(Q_m \in \mathbb{R}^{F_m \times K}\). The input to the LLM comprises the concatenation of \(Q_t\) and \(Q_m\) for \(m=\{v,s,o\}\), structured according to the template:
$[
\text{USER:} \hspace{0.05in} \langle Q_t \rangle \hspace{0.05in} \langle Q_m \rangle \hspace{0.05in} \text{Assistant:}
]
$
This stage 1 training ensures that the video, skeleton, and HOI cues are independently aligned to the LLM embedding space of \modelname.

\textbf{Sequence of Integrating Modalities.} Our approach to determining the sequence for integrating modalities leverages the principles of curriculum learning.Curriculum learning~\cite{MAVREC}, a machine learning strategy inspired by the pedagogical approach of progressing from easy to complex tasks, is utilized to structure the integration sequence in \modelname. MMPro training strategy adopts an incremental difficulty progression for multi-modal alignment. The complexity measure guiding the integration sequence is derived from the optimized autoregressive loss obtained from stage 1. Our findings suggest a gradation of integration difficulty among the modalities, with videos and skeletons being relatively simpler to align compared to HOI. The sparsity of HOI features, indicated by the infrequency of distinct HOIs within videos, explains the challenge of aligning HOI with the language embedding space. Consequently, in \modelname, the modalities are integrated in the order of skeletons followed by HOI, adhering to the curriculum of escalating difficulty.

In the second stage of training, \modelname's architecture expands to include additional modality-specific connector. These connector facilitate the simultaneous alignment of video and skeleton data with the LLM embedding. The parameters for this stage, $\theta_v$ and $\theta_s$, inherit their initial values from the weights optimized during stage 1. Consequently, the input format to the LLM is structured as follows:
$[
\text{USER:} \hspace{0.05in} \langle Q_t \rangle \hspace{0.05in} \langle Q_v \rangle \hspace{0.05in} \langle Q_s \rangle \hspace{0.05in} \text{Assistant:}
]
$
where $\langle Q_t \rangle$, $\langle Q_v \rangle$, and $\langle Q_s \rangle$ represent the text, video, and skeleton query embeddings, respectively. This structured input format ensures a targeted integration of video and skeleton modalities during the MMPro training strategy in stage 2.

\begin{figure}[htb]
    \centering
    \includegraphics[width=0.96\linewidth]{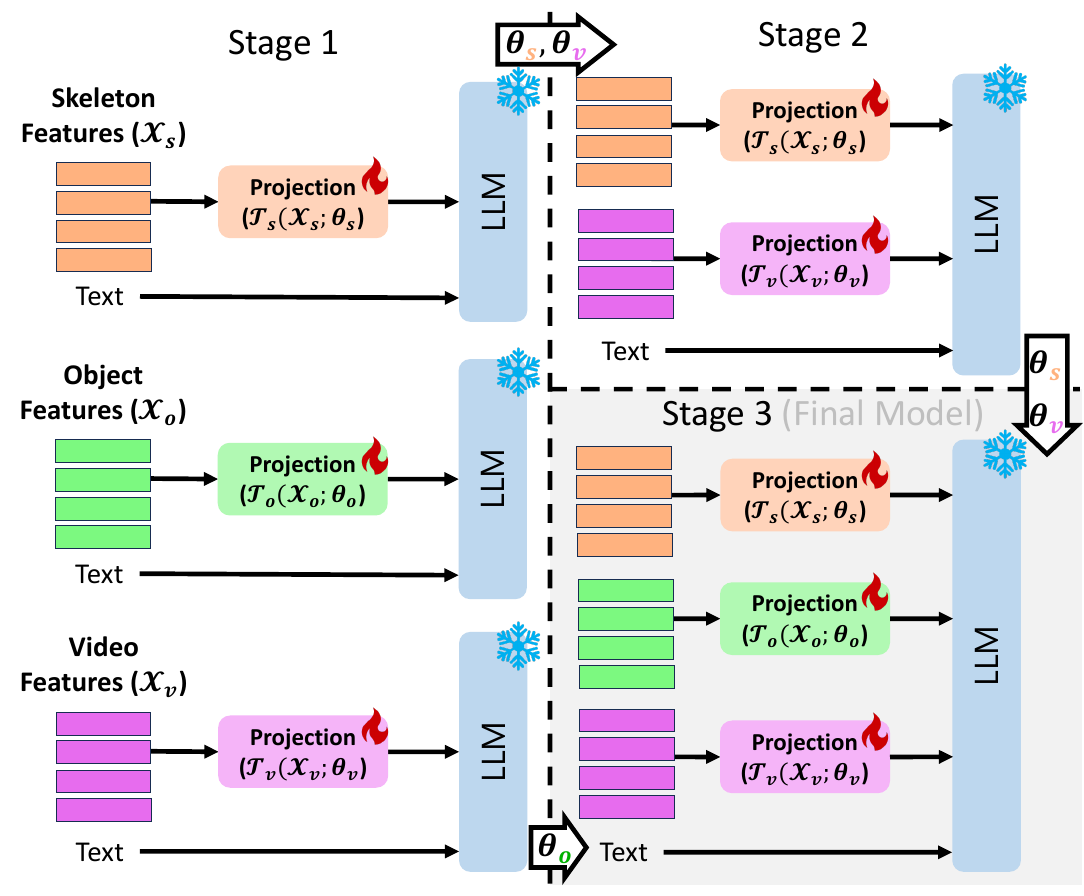}\vspace{-0.1in}
    \caption{\textbf{MMPro training.} Our proposed three-stage progressive training pipeline used to train \modelname. Stage 1 initializes independent projections for skeleton, object, and video features. Stage 2 combines skeleton and video modalities. Stage 3 integrates all modalities into the final model. Large hollow arrows indicate weight transfer between stages.}\vspace{-0.1in}
    \label{fig:mmpro}
    \vspace{-1em}
\end{figure}

The final integration stage in \modelname~incorporates all modalities. The training parameters $\theta_v$ and $\theta_s$ are further refined from their stage 2 configurations, while $\theta_o$ is initialized from stage 1 training. The input to the LLM at this stage includes an additional object modality, formatted as:
$[
\text{USER:} \hspace{0.05in} \langle Q_t \rangle \hspace{0.05in} \langle Q_v \rangle \hspace{0.05in} \langle Q_s \rangle \hspace{0.05in} \langle Q_o \rangle \hspace{0.05in} \text{Assistant:}
]
$.
This integration approach, as shown in Figure~\ref{fig:mmpro}, aligns video, object, and skeleton modalities with the LLM embeddings, enhancing the model's capability to accurately process and understand ADL.

When performing \textbf{inference}, \modelname~utilizes only the video cue, consequently eliminating the need for person-centric cropping and additional modalities.

\spacebeforesection
\section{Experiments}
\spaceaftersection
In this section we first discuss implementation details, then introduce the ADL related tasks used to evaluate state-of-the-art LLVMs, the evaluation metrics, and finally we present all of the results and analysis.

\spacebeforesection
\subsection{Implementation Details}
\spaceaftersection
During training, the embedding dimensions are $D_v = 1024$ for visual features extracted from CLIP-L/14~\cite{CLIP},  $D_o = 512$ for object features extracted from OWLv2~\cite{OWLv2}, $D_s = 216$ for skeleton features from SkeletonCLIP~\cite{sinha2025skimodels}. The embedding dimension of the Vicuna~\cite{vicuna2023} LLM is $K=4096$, and the number of visual and skeleton tokens are set as $F_v=356$ and $F_s=256$ respectively. The videos input to \modelname~ contain $T=100$ frames, with a spatial resolution of $H=224, W=224$. \modelname~ is trained on 8 NVIDIA RTX A6000 GPUs for $3$ epochs with a total batch size of $32$ and a learning rate of $2e^{-5}$. 
For the purpose of promoting research in this field, we also provide the extracted skeleton features and object trajectories of \modelname~along with the ADL-X dataset.

\spacebeforesection
\subsection{ADL Tasks, Test Datasets, and Metrics}
\spaceaftersection
This paper introduces novel benchmarks for assessing LLVM's temporal understanding of diverse and real-world ADL videos.
We propose two new \textbf{ADL MCQ} benchmarks including \underline{ADL MCQ-AR} and \underline{ADL MCQ-TC}. ADL MCQ-AR involves multiple-choice question-answering for action recognition, where the model selects the correct action from a set of options given a question about the action performed in a video. Similarly, ADL MCQ-TC focuses on predicting the missing action in a temporal sequence, given a video containing either a complete or incomplete sequence, including the action at the blank position (e.g., ``\textit{take something off table}, ``\textit{drink from cup}", $<$\texttt{blank}$>$, ``\textit{walk away}"). If the $<$\texttt{blank}$>$ appears at the end of an incomplete sequence where the input video lacks subsequent actions, the task is an action forecasting. Finally, we also evaluate the LLVMs on \textbf{ADL video description} tasks for long-duration videos with densely occurring actions and fine-grained captions.
Note that all our evaluations are performed zero-shot.

\textbf{Evaluation Datasets.} 
For ADL MCQ-AR evaluation, we utilize the Charades~\cite{charades} (Char) and Toyota Smarthome~\cite{smarthome} (SH) datasets. Evaluation for ADL MCQ-TC is conducted using LEMMA~\cite{lemma} (LEM) and Toyota Smarthome Untrimmed~\cite{tsu} (TSU) datasets. 
ADL MCQ-TC on LEMMA is an action forecasting task.
Video description tasks are assessed using the Charades and TSU datasets, both featuring long-duration videos with multiple actions per video. Notably, for the TSU dataset, we manually annotated video descriptions with fine-grained details regarding activities performed by elderly individuals, employing 6 human annotators for 174 videos. Our evaluation relies on these annotated descriptions, which we also provide to the community as part of the test set for \datasetname.

\textbf{Evaluation Metrics.} 
To ensure result consistency and reproducibility, we employ LLAMA 3.1~\cite{llama} locally instead of the GPT API, where model weights may update over time. For ADL MCQ task evaluation, we follow~\cite{mmbench, mvbench, liu2024tempcompass}, prompting LLAMA 3.1 with the output of \modelname, along with MCQ questions and options, to select the nearest option generated by \modelname, thus minimizing scoring bias. For video-level description evaluation, following~\cite{videochatgpt}, generated descriptions are compared against ground truth and scored by LLAMA across dimensions of Information Correctness, Detail Orientation, Contextual Understanding, Temporal Understanding, and Consistency, with scores scaled to a maximum of 100.



\begin{table}[t]
\centering
\caption{Impact of different components in the \datasetname~data curation framework on LLVM training.}
\resizebox{0.94\linewidth}{!}{
\begin{tabular}{lccccccc}
\toprule
\multirow{2}{*}{\textbf{Training data}} & \multirow{2}{*}{\colorbox{green}{\textbf{PAG}}} & \multirow{2}{*}{\colorbox{pink}{\textbf{TS}}} & \multirow{2}{*}{\colorbox{chrome-yellow}{\textbf{WS}}} & \textbf{ADL MCQ-AR} & \textbf{ADL MCQ-TC} & \textbf{Descrip. (Avg)} \\
& & & & \textbf{SH} & \textbf{TSU} & \textbf{TSU}\\
\midrule
ActivityNet & \ding{55} & \ding{55} & \ding{55} & 39.6 & 20.9 & 31.9 \\
NTU120 & \checkmark & \ding{55} & \checkmark & 37.2 & 20.5 & 54.1 \\
ADL-X & \ding{55} & \checkmark & \checkmark & 40.6 & 25.5 & 59.1 \\
ADL-X & \checkmark & \checkmark & \ding{55} & 39.0 & 24.8 & 62.2 \\
\rowcolor{LightBlue}
ADL-X & \checkmark & \checkmark & \checkmark & \textbf{44.5} & \textbf{29.5} & \textbf{64.8} \\
\bottomrule
\end{tabular}}
\label{impact_dataset}
\vspace{-1.2em}
\end{table}

\spacebeforesection
\subsection{Impact of \datasetname~Training on LLVMs}
\spaceaftersection
In Table~\ref{impact_dataset}, we demonstrate the effectiveness of our novel semi-automated curation of the \datasetname~dataset. 
To validate the importance of domain-specific training, we trained an LLVM on trimmed NTU120 clips, showing that the NTU120-trained model outperforms the model trained on ActivityNet~\cite{caba2015activitynet} for description task, while our \datasetname-trained LLVM achieves the highest performance in all ADL tasks. 
As shown in Table~\ref{impact_dataset}, all strategies—\colorbox{green}{PAG}, \colorbox{pink}{TS}, and \colorbox{chrome-yellow}{WS}—used to curate \datasetname~from NTU120 generate a high-quality video-instruction pairs with minimal hallucinations, enabling the LLVM to effectively understand human-centered actions and handle temporal randomness.

\textbf{Verification of \datasetname.} To validate the correctness of our LLM-assisted data curation pipeline, we conducted a human evaluation study where the quality of 100 video-QA pairs were rated on a scale of 1-5 based on QA pair correctness. This user study revealed an average rating of 4.1 for the the generated QA pairs, demonstrating that our three key filtering strategies successfully yields high quality video-instruction pairs.

\begin{table}[t]
\centering
\caption{Strategies to integrate skeletons and HOIs in LLVM.}
\setlength{\tabcolsep}{3.5pt}
\footnotesize
\resizebox{0.94\linewidth}{!}{
\begin{tabular}{lccccc}
\toprule
\multirow{2}{*}{\textbf{Method}} & \multicolumn{2}{c}{\textbf{ADL MCQ-AR}} & \multicolumn{2}{c}{\textbf{ADL MCQ-TC}} & \textbf{TSU} \\
& \textbf{Char} & \textbf{SH} & \textbf{LEM} & \textbf{TSU} & \textbf{Descrip. (Avg)} \\
\midrule
\datasetname~ChatGPT & 51.0 & 44.5 & 28.6 & 29.5 & 64.8 \\
\midrule
\multicolumn{6}{c}{\cellcolor{gray!20}\hspace{0.5in}\textit{Video+Skeleton}} \\
Skeleton Features (SF) & \textbf{52.7} & 42.6 & 33.1 & 30.3 & 66.5 \\
Skeleton QA & 47.5 & 43.1 & 25.9 & 29.6 & 61.8 \\
Skeleton Context (SC) & 48.2 & 45.4 & 27.8 & 30.2 & 66.3 \\
\rowcolor{LightBlue}SC + SF & 51.2 & \textbf{46.2} & \textbf{33.5} & \textbf{32.5} & \textbf{66.8} \\
\midrule
\multicolumn{6}{c}{\cellcolor{gray!20}\hspace{0.5in}\textit{Video+HOI}} \\
\rowcolor{LightBlue}HOI Features & \textbf{53.8} & \textbf{48.0} & 32.6 & \textbf{37.1} & \textbf{68.0} \\
HOI (YOLO+CLIP) & 53.7 & 45.1 & \textbf{34.1} & 37.0 & 67.8 \\
HOI QA & 50.4 & 45.5 & 30.3 & 28.4 & 62.5 \\
HOI Context & 50.3 & 44.6 & 26.5 & 27.8 & 63.8 \\
\bottomrule
\end{tabular}}\vspace{-0.1in}
\label{tab:features}
\end{table}

\begin{table}
\centering
\caption{Integrating all modalities in \modelname.}\vspace{-0.1in}
\setlength{\tabcolsep}{3pt}
\footnotesize
\resizebox{0.94\linewidth}{!}{
\begin{tabular}{lccccc}
\toprule
\multirow{2}{*}{\textbf{Method}} & \multicolumn{2}{c}{\textbf{ADL MCQ-AR}} & \multicolumn{2}{c}{\textbf{ADL MCQ-TC}} & \textbf{TSU} \\
& \textbf{Char} & \textbf{SH} & \textbf{LEM} & \textbf{TSU} & \textbf{Descrip. (Avg)} \\
\midrule
SF + OF & 53.8 & 40.7 & 32.1 & 33.1 & 65.8 \\
X-InstructBLIP~\cite{panagopoulou2023xinstructblip} & 49.0 & 45.6 & 27.5 & 29.9 & 65.5 \\
\midrule
\multicolumn{6}{c}{\hspace{0.5in}\cellcolor{gray!20}\textit{MMPro Training}} \\
\rowcolor{LightBlue}
Prog. A (Token) & \textbf{55.2} & 48.1 & \textbf{34.3} & \textbf{38.2} & \textbf{70.8} \\
Prog. A (String) & 52.7 & 45.4 & 32.3 & 34.5 & 65.5 \\
Prog. B (Token) & 52.8 & \textbf{49.4} & 32.8 & 33.0 & 69.2 \\
Prog. B (String) & 51.3 & 48.6 & 30.1 & 32.6 & 67.4 \\
Prog. A (SF+SC, Token) & 54.5 & 49.3 & 32.2 & 34.9 & 69.5 \\
\bottomrule
\end{tabular}
}
\label{tab:progressive}
\vspace{-1em}
\end{table}

\begin{table*}[t]
\caption{\textbf{State-of-the-art comparison.} Performance is shown on ADL MCQ and ADL Video Description tasks. [CI: \textit{Correctness of Information}, DO: \textit{Detail Orientation}, CU: \textit{Contextual Understanding}, TU: \textit{Temporal Understanding}, Con: \textit{Consistency}]}\vspace{-0.1in}
\centering
\resizebox{0.94\linewidth}{!}{
\begin{tabular}{lc|cc|cc|ccccc|c|ccccc|c}
\toprule
\multirow{2}{*}{\textbf{Method}} & \textbf{Training} & \multicolumn{2}{c|}{\textbf{ADL MCQ-AR}} &\multicolumn{2}{c|}{\textbf{ADL MCQ-TC}} & \multicolumn{5}{c|}{\textbf{Charades}} & \multirow{2}{*}{\textbf{Avg}} & \multicolumn{5}{c|}{\textbf{TSU}} & \multirow{2}{*}{\textbf{Avg}}\\
& \textbf{Data Size} & \textbf{Char} & \textbf{SH} & \textbf{LEM} & \textbf{TSU} & \textbf{CI} & \textbf{DO} & \textbf{CU} & \textbf{TU} & \textbf{Con} & & \textbf{CI} & \textbf{DO} & \textbf{CU} & \textbf{TU} & \textbf{Con} & \\
\midrule
CogVLM~\cite{cogvlm} + GPT~\cite{gpt} & 1.5B Images & 52.3 & 42.5 & 32.0 & 23.6 & 42 & 62 & 49.6 & \textbf{36.5} & 32.8 & 44.6 & 55.2 & 72.0 & 60.6 & 30.2 & 48.5 & 53.3\\
CogVLM~\cite{cogvlm} + Llama~\cite{llama} & 1.5B Images & 52.8 & 43.2 & 32.5 & 22.5 & 40.2 & 61.8 & 49.5 & \textbf{36.5} & 33.5 & 44.3 & 49.8 & 66 & 56.6 & 29.8 & 40.2 & 48.5\\
BLIP2~\cite{blip2} + GPT~\cite{gpt} & 1.5B Images  & 50.2 & 39.6 & 28.9 & 20.2 & 39.8 & 60.2 & 47.8 & 36.0 & 37.2 & 44.2 & 48.8 & 66.6 & 63.6 & 45.6 & 39.8 & 52.9\\
\midrule
VideoLlama~\cite{videollama} & 2.6M QA Pairs  & 40.2 & 44.8 & 32.6 & 24.6 & 22.2 & 42.5 & 33.8 & 20.2 & 34.5 & 30.6 & 57.8 & 62.0 & 62.4 & 48.2 & 44.4 & 54.9\\
VideoLlava~\cite{videollava} & 1.2M QA Pairs  & 41.8 & \textbf{49.2} & 30.0 & 25.5 & 23.6 & 46.4 & 34 & 20.6 & 33.5 & 31.6 & 30.8 & 54.8 & 42.4 & 30.4 & 44.5 & 40.6\\
VideoChatGPT~\cite{videochatgpt} & 100K QA Pairs  & 51.0 & 39.6 & 31.4 & 20.9 & 26.1 & 45.2 & 35.6 & 21.4 & 31.2 & 31.9 & 31.2 & 52.8 & 78.2 & \textbf{64.8} & 45.6 & 54.5\\
ChatUniVi~\cite{jin2023_chatunivi} & 3M QA Pairs  & 53.1 & 48.1 & 32.3 & 36.4 & 36.5 & 54.5 & 46.6 & 32.2 & 35.9 & 41.1 & 56.8 & 66.9 & 79.0 & 50.0 & 56.6 & 61.9 \\
\midrule
\datasetname~ChatGPT~\cite{videochatgpt} & 100K QA Pairs & 51.0 & 44.5 & 28.6 & 29.5 & 40.6 & 50.6 & 49.8 & 30.6 & \textbf{40.2} & 42.4 & 62.4 & 79.4 & 70.8 & 51.2 & 60.4 & 64.8\\
\rowcolor{LightBlue} \textbf{\modelname~} & 100K QA Pairs & \textbf{55.2} & 48.1 & \textbf{34.3} & \textbf{38.2} & \textbf{45.8} & \textbf{64.2} & \textbf{57.0}  & 36.4  & 39.4 & \textbf{48.6} & \textbf{66.0}  & \textbf{86.2} & \textbf{79.6} & 50.0 & \textbf{72.4} & \textbf{70.8}\\
\bottomrule
\end{tabular}}
\vspace{-0.2in}
\label{tab:sota}
\end{table*}

\begin{figure}
    \centering
    \includegraphics[width=\linewidth,  height=9cm]{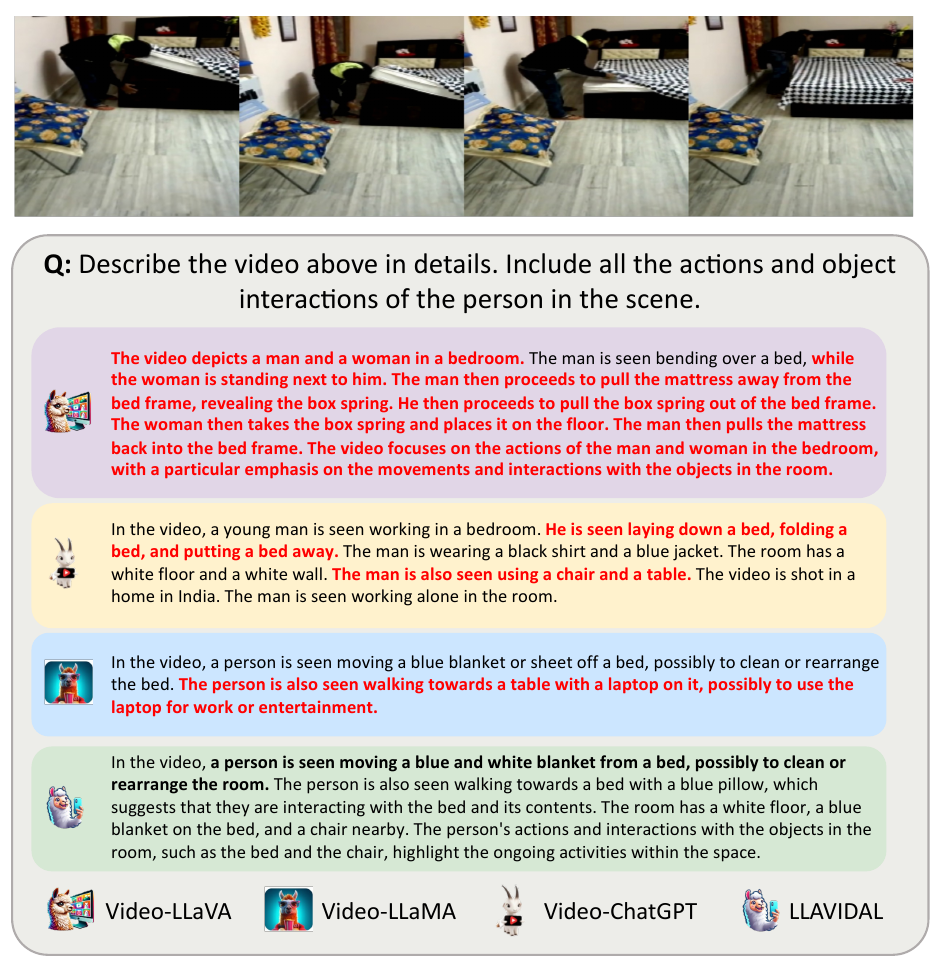}
    \vspace{-1.8em}
    \caption{\textbf{Qualitative results comparing \modelname~with SOTA models.} \textbf{\textcolor{red}{Incorrect}} descriptions are marked in \textbf{\textcolor{red}{red}}.}
    \label{fig:visualization}
    \vspace{-1.8em}
\end{figure}

\spacebeforesection
\subsection{How to integrate modalities in LLVMs?}
\spaceaftersection

Table~\ref{tab:features} explores the integration of skeleton and HOI cues alongside RGB videos in \modelname. We integrate skeleton information as features (SF), QA pairs, and context (SC) as illustrated in section~\ref{multimodal}. Both SF and SC outperform the \datasetname~ChatGPT baseline (Video ChatGPT model trained with \datasetname) across most ADL tasks, with the combination of skeleton context and features (SC+SF) achieving the best results, corroborating the importance of skeleton integration for enhanced ADL understanding in LLVMs.
In contrast, HOI cues used as QA or context provide limited discriminative value for the LLM. However, HOI features extracted from ObjectLM significantly improve performance across ADL tasks, showing their crucial role in ADL understanding. We find that HOI extraction using BLIP2 + OWLv2 outperforms the conventional object detector + CLIP approach, supporting our choice for extracting HOI. 

Table~\ref{tab:progressive} shows the performance of \modelname~using various strategies to jointly integrate modalities $\mathcal{M}_m$, namely videos, skeletons, HOIs, and language. Joint training of all modalities and modality-specific alignment with language following X-InstructBLIP\cite{panagopoulou2023xinstructblip} underperforms the model trained in Table~\ref{tab:features} due to difficulties in optimizing the projection layer $\mathcal{T}_v$ for effective alignment of both $\mathcal{T}_p$ and $\mathcal{T}_o$. 
In contrast, our \textit{MMPro} strategy effectively retains discriminative representations from all modalities through its stage-wise integration—first video, then skeletons, and finally HOIs—which addresses optimization challenges faced by baseline multimodal models. To validate the modality integration sequence in \modelname, we conduct experiments with a curriculum-based model (Prog. A) using stage 1 MMPro losses and an alternative integration sequence—video, then objects, then skeletons (Prog. B). We also ablate the impact of prefixing each modality with a token or string. Results show that Prog. A with token-type prefixing per modality performs best across most ADL tasks, further confirming the efficacy of curriculum-based modality integration in MMPro. 
Moreover, we find that skeleton features alone are sufficient, and skeleton context is not needed when integrating skeletons in \modelname.

\spacebeforesection
\subsection{Comparison to the state-of-the-art}
\spaceaftersection
We compare \modelname~against the state-of-the-art on ADL MCQ tasks and ADL video description tasks.

\textbf{ADL MCQ.} Table~\ref{tab:sota} compares \modelname~to state-of-the-art LLVMs on the ADL MCQ-AR and ADL MCQ-TC benchmarks. \modelname~achieves substantial improvements in action recognition, outperforming the representative baseline, VideoChatGPT, by \textcolor{codegreen}{+8.2\%} and \textcolor{codegreen}{+21.4\%} on the Charades and Smarthome datasets, respectively. On the TSU dataset, \modelname~surpasses VideoChatGPT by up to \textcolor{codegreen}{+82.7\%}, highlighting its superior temporal understanding.
Figure~\ref{fig:visualization} provides a visual comparison of \modelname~against representative baselines on the ADL benchmarks, with additional examples in Appendix~\ref{sec:qual_eval}.

\textbf{ADL Video Description.} 
Table~\ref{tab:sota} shows the performance comparison of baseline LLVMs and \modelname~on their video description capabilities on the Charades and TSU datasets. 
Video-level descriptions are obtained directly from the Charades dataset. For the TSU dataset, comprising lengthy videos, we segment each video into 1-minute clips and input them individually to the LLVMs for generating clip-level descriptions. Subsequently, we concatenate all clip-level descriptions and utilize GPT-3.5 turbo to summarize them into a  video-level description, following the same instruction template utilized in our dense description pipeline for \datasetname. Across all the 5 VideoChatGPT metrics, the results show that
\modelname~consistently surpasses current SOTA methods, and outperforms all models including the image captioners-summarizers pipelines which are trained on billions of images.

\spacebeforesection
\section{Conclusion}
\spaceaftersection
In this work, we present a novel semi-automated framework for curating ADL video instruction tuning dataset and introduce \datasetname. 
We introduce \modelname, an LLVM capable of integrating 3D skeletons and HOI modalities using a novel Multimodal Progressive training strategy.  
To assess LLVMs' performance in ADL scenarios, we propose the ADL MCQ and video description benchmarks. Results demonstrate that \modelname, when trained on \datasetname, surpasses other baselines, indicating its efficacy in grasping intricate temporal relationships within ADL contexts. 

\section*{Acknowledgments}
This work is supported in part by the National Science Foundation (IIS-2245652) and NSF's NAIRR Pilot initiative (NAIRR240338), which provided credits to access GPT-3.5 Turbo. This work was also supported by funds provided by The University of North Carolina at Charlotte.

{
    \small
    \bibliographystyle{ieeenat_fullname}
    \bibliography{refs,main}
}

\maketitlesupplementary
\appendix
\section{Overview}

The Supplementary material is organized as follows:

\begin{itemize}
    \item Section \ref{sec:rel-work}: Related Work
    \item Section \ref{sec:data}: Additional Dataset Details
    \item Section \ref{sec:implement}: Additional Implementation Details
    \item Section \ref{sec:actions}: Improving Actions: Skeleton Cues vs Object Cues
    \item Section \ref{sec:qual_eval}: Additional Qualitative Evaluation
    \item Section \ref{sec:prompts}: LLM Prompts Used
    \item Section \ref{sec:limit}: Limitations
    \item Section \ref{sec:license}: Licensing  and Intended Use

\end{itemize}


\section{Related Work}
\label{sec:rel-work}
In this section, we discuss the recent datasets proposed for instruction tuning of LLVMs. We also present the recent advancements in multimodal conversational models both with training-free methods and methods using visual connectors.

\begin{table*}[t]
\centering
\caption{Video Instruction Dataset Comparison.}
\begin{adjustbox}{max width=\textwidth}
\scalebox{0.8}{
\begin{tabular}{lccccccccc}
\toprule
\textbf{Dataset} & \textbf{Modalities} & \textbf{Subjects} & \textbf{Multiple} & \textbf{Videos} & \textbf{QA Pairs} & \textbf{Atomic Actions}  & \textbf{Temporal }  & \textbf{Object} & \textbf{Type}\\
 & & & \textbf{Views} & & & \textbf{per Vid} & \textbf{Rand.} & \textbf{Traj.} & \\
\midrule
TimeIT\cite{timechat} & RGB+L & NA & No & 173000 & 173K & Medium & No & No & Web \\
VideoChat\cite{videochat} & RGB+L & NA & No & 8196 & 11K & Low & No & No & Web \\
Valley\cite{luo2023valley} & RGB+L & NA & No & 64,687 & 65K & Low & No & No & Web \\
VideoChatGPT~\cite{videochatgpt} & RGB+L & NA & No & 27,801 & 100K & Medium & No & No & Web \\
\midrule
\rowcolor{gray!25}\datasetname & RGB+S+L & 106 & Yes & 16,343 & 100K & High & Yes & Yes & ADL \\
\bottomrule
\end{tabular}}
\end{adjustbox}
\label{tab:data_comp}
\end{table*}

\textbf{Data:} Existing video-instruction datasets, such as VideoChat\cite{videochat}, Valley\cite{luo2023valley}, Video-ChatGPT~\cite{videochatgpt}, and TimeIT~\cite{timechat}, have made significant strides in advancing general video understanding and dialogue. 
Valley is derived from a website called Jukinmedia that provides videos with diverse categories and wide detailed descriptions.
TimeIT dataset from TimeChat offers videos with temporal variations and task diversity.
On the other hand, ActivityNet~\cite{caba2015activitynet} boasts a diverse taxonomy with 203 activity classes, most activity classes are not tailored to the ADL domain. It is to be noted that most LLVMs like VideoChatgpt~\cite{videochatgpt}, VideoLlava~\cite{videollava} derive their instruction dataset from ActivityNet. Webvid, which is now de-comissioned due to privacy issues introduced in~\cite{frozen_in_time}, consists of 2.5 million video-text pairs scraped from the web.
Although these video instruction datasets are large in scale, offer diverse action classes, and focus on general video understanding, they fail to address the challenges specific to ADL. These challenges include intra-class temporal variations, long-term temporal relationships, complex human-object interactions, and videos captured in multiview settings. Unlike ADL videos, the internet videos in these datasets are predominantly shot by a cameraman, ensuring human-centered frames. Consequently, they do not capture the unstructured randomness spatially and temporally inherent in real-world ADL videos. In contrast, \datasetname~is specifically designed to address the challenges inherent in ADL (see Table~\ref{tab:data_comp}). It captures temporal unstructuredness through the temporal stitching of several unrelated actions in sequence and incorporates complex human-object interactions from the NTU120 dataset. Additionally, our proposed PAG and WS video description techniques effectively eliminate hallucinations, resulting in high-quality video-instruction pairs.

\textbf{Image captioners + LLM}.
Advancements in Large Language Models (LLMs) have naturally extended vision-language models~\cite{CLIP}, leveraging LLMs to enhance reasoning capabilities. The emergence of these foundation models has given rise to training-free methods like Socratic Models~\cite{socratic} and VideoChat~\cite{videochat}, which use pretrained vision encoders~\cite{UniFormer, internvideo} to map visual information into a language embedding space, followed by LLMs for downstream video tasks.

On the other hand, effective image captioners like CogVLM~\cite{cogvlm} introduce separate layers into the Transformer block of the LLM to process image features using independent QKV matrices and Feed Forward Networks specifically designed for images. Such effective captioning approaches have inspired methods that map visual information to language via image captioners, followed by processing with LLMs.
Among dialog-based models, VideoChatCaptioner~\cite{ChatCaptioner} summarizes videos through conversations between ChatGPT~\cite{gpt} and a captioner such as BLIP2~\cite{blip2}. Similarly, ChatVideo~\cite{ChatVideo} employs task-specific foundation models to create a database of "tracklets," with a database manager and ChatGPT~\cite{gpt} collaborating to generate responses for user queries during inference. 

For long video understanding, approaches like \cite{llovi, LongVLM} segment videos into smaller units, providing either segment-level descriptions directly to LLMs or encoding each segment, concatenating tokens, and projecting them into the LLM space. Likewise, Language Repository~\cite{Kahatapitiya2024langrepo} introduces write-and-read operations to prune text redundancies and extract information across various temporal scales.
The Multimodal Video Understanding Framework~\cite{kanchana_onepass} explores integrating video-specific information into an LLM-based framework by using off-the-shelf vision tools to extract three object-centric modalities from videos and fusing this information through natural language. Additionally, \cite{Park2024TooMF} investigates optimal strategies for key-frame selection to significantly reduce redundancies.
However, despite these advances, training-free models fail to capture the complex temporal relationships intrinsic to ADL. These relationships, including long-term dependencies and intricate human-object interactions, remain a significant challenge for these approaches.

\textbf{Large Language Vision Models (LLVMs)}.
The abilities of LLMs in contextual understanding and language generation have led to the introduction of video conversational models. These methods typically employ foundation models to extract visual features from images and project them into an embedding space compatible with language models.  
Flamingo~\cite{flamingo} utilizes a vision-language resampler combined with gated cross-attention, while BLIP2~\cite{blip2} introduces Q-Former to map image features into the LLM embedding space. Similarly, MiniGPT4~\cite{minigpt} uses a simple linear projection layer. However, these models fall short of becoming conversational assistants due to the lack of human instruction feedback.  
To address this, mPLUG-OWL~\cite{mplugowl} first aligns visual and language features through multimodal autoregressive pretraining, followed by multimodal instruction tuning using LoRA~\cite{lora}, enabling more natural and human-aligned responses. Models such as InstructBLIP~\cite{instructblip} and LLaVA~\cite{llava} introduce large-scale human instruction datasets to facilitate LLM fine-tuning. Meanwhile, models like PaLI~\cite{pali} and Qwen-VL~\cite{qwen} allow direct training of LLMs during pretraining or supervised fine-tuning stages.  

Other models, including VideoChat, VideoLLaMA, and TimeChat~\cite{videochat, videollama, timechat}, leverage Q-Former for effective feature encoding and alignment. For example, VideoLLaMA~\cite{videollama} employs a vision transformer with an image Q-Former to obtain frame-level representations, followed by a video Q-Former for temporal modeling. Similarly, TimeChat~\cite{timechat} encodes variable-length videos using a timestamp-aware frame encoder with a Q-Former to infuse temporal information into vision tokens, followed by a sliding window Q-Former to condense frame-level features for the projection layer.  
Building on these approaches, VideoLLaVA~\cite{videollava} jointly trains on both images and videos, pre-aligning visual modalities to language using LanguageBind~\cite{LanguageBind} encoders. VideoChatGPT~\cite{videochatgpt} leverages both temporal and spatial features from videos, obtained by average pooling frame-level features both spatially and temporally.  

In contrast to these models, \modelname~incorporates 3D skeleton data and human-object interaction (HOI) cues alongside videos into the LLM embedding space. This integration enables the additional cues to learn discriminative video representations, making \modelname~particularly effective for interpreting ADL videos, where temporal relationships and complex HOIs are crucial.

\begin{figure*}[h]
    \centering
    \includegraphics[width=\textwidth]{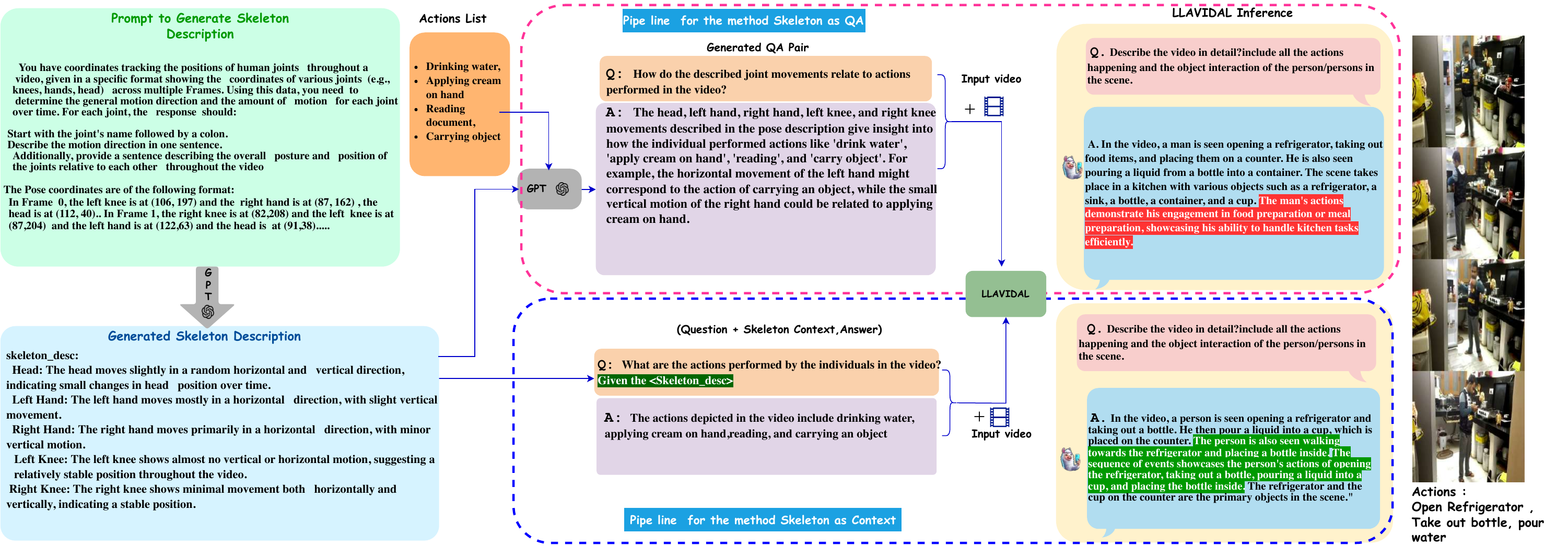}
    \caption{Overview of pipeline for Skeleton as QA and Skeleton as context.}
    \label{fig:skeleton_qacon_pipeline}
\end{figure*}
\begin{figure*}[h]
    \centering
    \includegraphics[width=\textwidth]{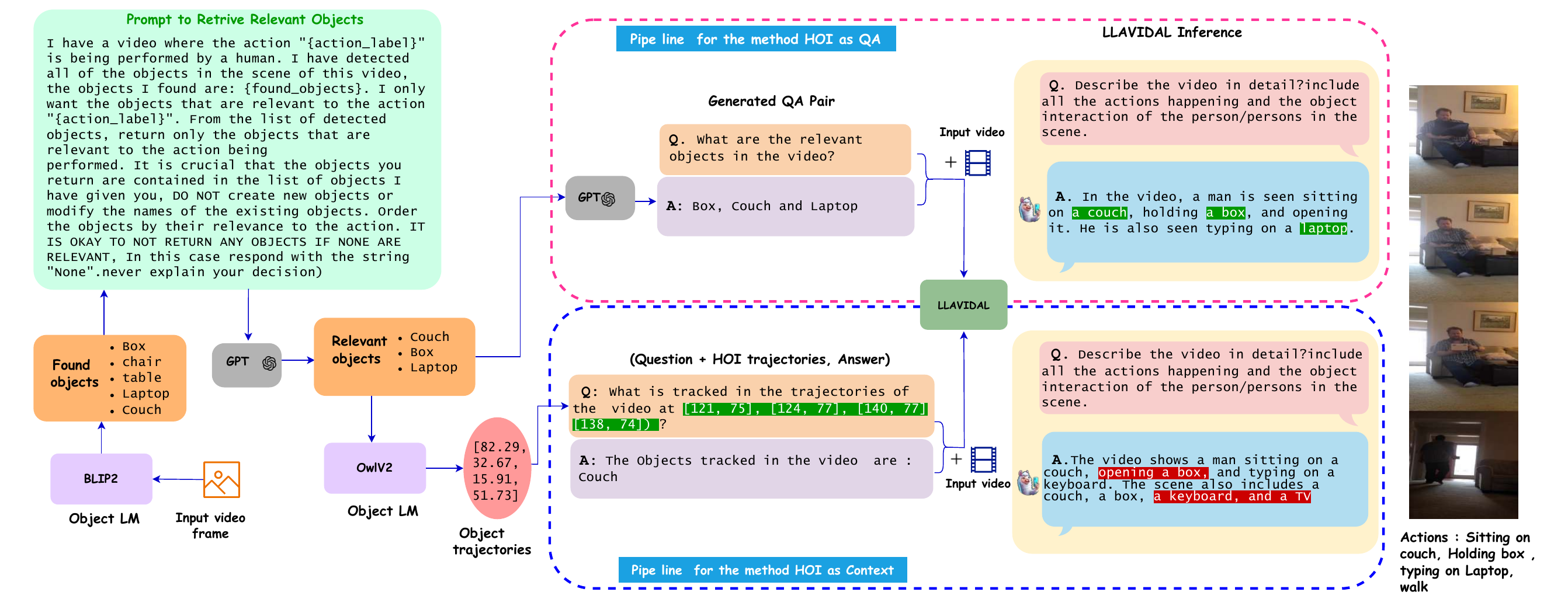}
    
    \caption{Overview of pipeline for HOI as QA and HOI as context.}
    \label{fig:object_qacon_pipeline}
\end{figure*}

\section{Additional Dataset Details}
\label{sec:data}
\textbf{Question Types.} We divide our QA in different questions so that our model understands human object interaction holistically, we lay emphasis on actions performed and the sequence of actions occurring in the video and likewise how objects are associated with the actions. We carefully design such questions relevant to the videos with GPT 3.5 Turbo. The questions encompasses \textit{actions happening, summarization, objects in the scene, color of the objects and questions related to the video}. For Skeleton as QA and Object as QA, we construct two additional questions for each. For Skeleton, we include \textit{"What is the motion of the body and joints relative to the actions?"} and \textit{"Which joints are moving in the video?"}. For Object, we add \textit{"What are the relevant objects in the scene?"} and \textit{"What is the object in the trajectory [x1, y1, x2, y2]?"}. These are illustrated in Figures~\ref{fig:skeleton_qacon_pipeline} and~\ref{fig:object_qacon_pipeline}.

\begin{figure*}[h]
    \centering
    \includegraphics[width=\textwidth]{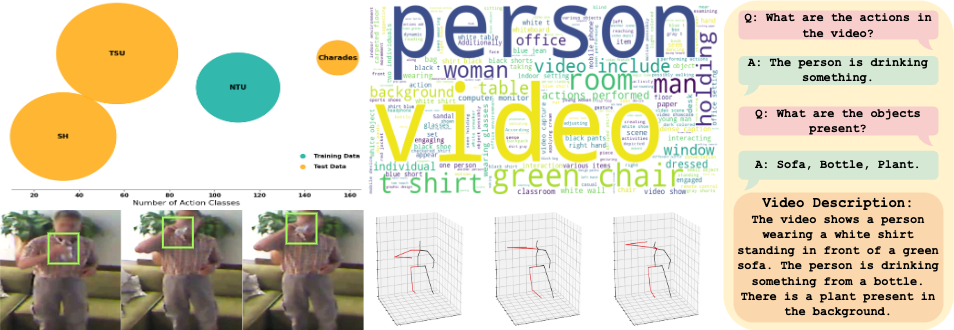}
    \caption{Overview of \datasetname. Top Left: Training and test data distribution; Top Middle: Wordcloud of Textual Representation of Training Data; Bottom Left: Sample video frames with detected relevant object Bottom Middle: 3D skeletons of the corresponding sample video; Right: Sample QA pairs}
    \label{fig:dataset_overview}
\end{figure*}

\textbf{Average video and sentence length.} There is an average of $23$ words per sentence in our QA and average word count for each answer is $42$. The average video length is $10$ seconds in our dataset. We have $126,2229$ nouns,
 $551,172$ verbs, $40,415$ actions and $722,807$ objects in our QA showing the overall dynamics of dataset which is illustrated in WordCloud of the Figure~\ref{fig:dataset_overview}.

\begin{figure*}[h]
    \centering
    \includegraphics[width=\textwidth]{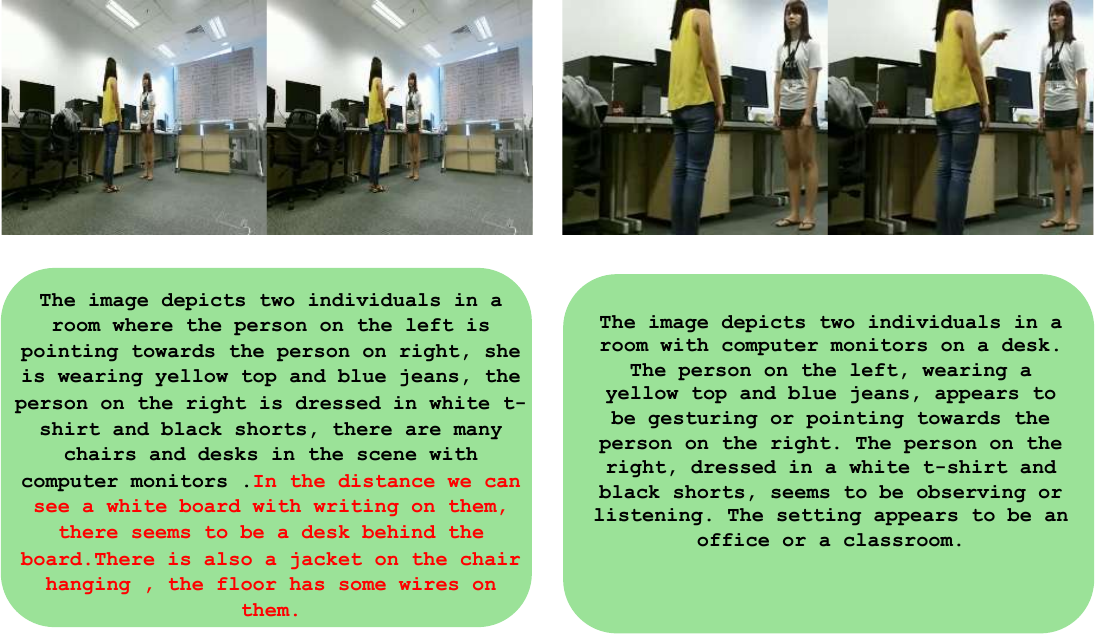}
    \caption{Left: uncropped videos and frame level annotations from CogVLM; Right: PAG and CogVLM captions. The irrelevant information (marked \textcolor{red}{red}) adds noise to the annotations.}
    \label{fig:uncroppedvscropped}
\end{figure*}

\textbf{Importance of Person Augmented Generation (PAG).}
When PAG is not applied to the videos in an ADL dataset, the resulting dense-level captions often include a significant amount of irrelevant information about the background scene. This extraneous information is not directly connected to the subject or the actions being performed, and its presence can introduce noise into the training data. If left unchecked, this noise can have a detrimental effect on the learning process, as the model may erroneously focus on the background details rather than the key elements of the ADL. By failing to isolate the relevant information, the model's attention is diverted away from the crucial aspects of the task at hand, namely the individual performing the actions, the actions themselves, and the interactions between the person and objects in the scene. This dilution of focus can lead to suboptimal performance and hinder the model's ability to accurately understand and classify ADLs.  In contrast, by employing person-centric cropping, the irrelevant background information is effectively eliminated from the videos. This targeted approach ensures that the dense-level captions concentrate solely on the elements that are directly related to the subject and their actions. By maintaining this persistent focus on the relevant information, the training data becomes more coherent and informative, enabling the model to better capture the essential characteristics of the ADLs. In Fig~\ref{fig:uncroppedvscropped}, we illustrate an example to highlight the importance of PAG in our semi-automated data curation framework.


\section{Additional Implementation Details}
\label{sec:implement}
We deployed a 4-bit quantized version of CogVLM-17B~\cite{cogvlm} for annotating frame-level captions. On an A5000 GPU, the inference uses 11GB of memory.The two prompts that are used to get the frame-level descriptions for the \datasetname~are -- \textit{"Give a detailed description of the actions happening and describe the image, include motions and the objects interacted by the person"} and \textit{"Summarize the content of the image in details explaining all events happening"}. CogVLM uses Vicuna v1.5 7b~\cite{vicuna2023} as their large language model and EVA2-CLIP-E~\cite{eva_CLIP} as their VIT encoder, the input image dimensions are $224 \times 224$, the average time to annotate a video is \textit{80 seconds} at \textit{0.5fps}.

\textbf{\modelname~details.} To generate HOI cues, we perform frame-level object detection using BLIP2 and localization using OWLv2. BLIP2~\cite{blip2} uses a ViT-L and a FlanT5~\cite{Flan} architecture for detection, while OWLv2~\cite{OWLv2} uses an OWL-ViT-L which is a CLIP based model for extracting localization features of the detected objects. In case of skeletonCLIP, the skeleton Encoder, Hyperformer, is pretrained on NTURGBD for 140 epochs for action recognition, and then is aligned with the CLIP Text Encoder for an additional 100 epochs. 
\modelname~uses a Vicuna-v1.1 (7B) as the LLM which is frozen during instruction tuning. 



\section{Improving Actions: Skeleton vs HOI Cues}
 \label{sec:actions}

\begin{figure*}[h]
    \centering
    \includegraphics[width=\textwidth]{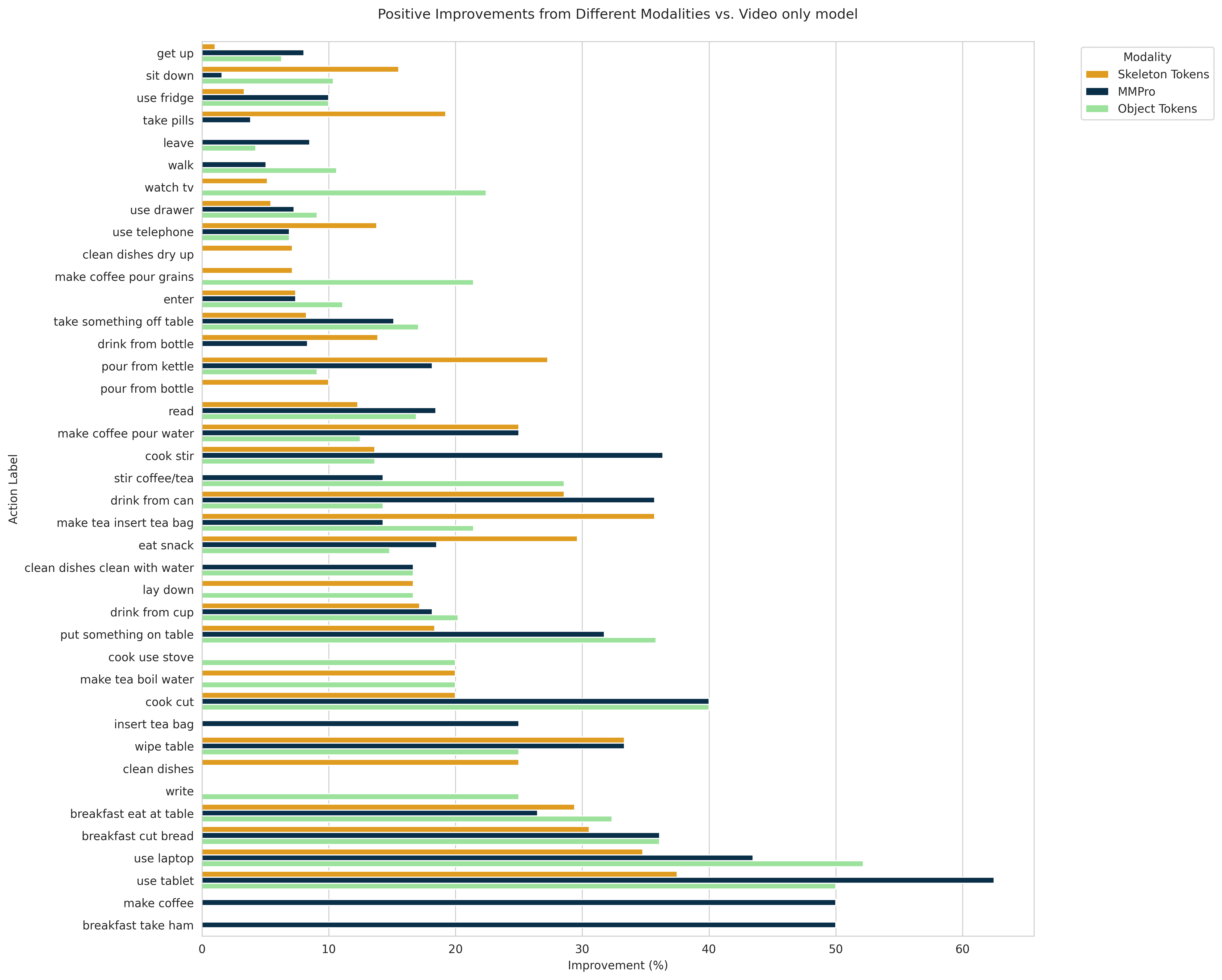} \vspace{-0.2in}
    \caption{Improvement of actions with object tokens vs skeleton tokens vs MMPro training}
    \label{fig:Pose-object-analysis}
\end{figure*}

In this section, we present a comprehensive analysis of our multi-modal approach leveraging HOI tokens, skeleton tokens, and their progressive integration (MMPro) for fine-grained action recognition. Our empirical evaluation across a diverse set of household actions demonstrates distinct performance patterns across modalities, revealing their complementary nature in action understanding when compared against video only trained model. This analysis is shown in Figure \ref{fig:Pose-object-analysis}.

HOI token integration along with video token in LLMs yields substantial performance gains in object-interaction intensive actions, with peak improvements in ``use laptop'' (\textcolor{OliveGreen}{+45\%}), ``use tablet'' (\textcolor{OliveGreen}{+32\%}), ``cook cut'' (\textcolor{OliveGreen}{+40\%}), and ``make coffee'' (\textcolor{OliveGreen}{+35\%}). This performance boost can be attributed to the enhanced object-contextual reasoning capabilities, where the model effectively leverages object-centric features to disambiguate actions occurring in similar spatial contexts. The HOI token cue demonstrates particular efficacy in scenarios requiring fine-grained object state understanding, such as distinguishing between ``pour from kettle'' (\textcolor{OliveGreen}{+20\%}), ``pour from can'' (\textcolor{OliveGreen}{+18\%}), and ``pour from bottle'' (\textcolor{OliveGreen}{+15\%}), where object state transitions and object-specific attributes are crucial for action classification. Additionally, HOI tokens show significant improvements in context dependent actions like ``breakfast eat at table'' (\textcolor{OliveGreen}{+25\%}), ``put something on table'' (\textcolor{OliveGreen}{+22\%}), and ``clean dishes dry up'' (\textcolor{OliveGreen}{+18\%}), where spatial relationships between multiple objects exists.

The skeleton token cue exhibits superior performance in actions characterized by distinctive kinematic patterns, showing significant improvements in body-centric actions such as ``take pills'' (\textcolor{OliveGreen}{+28\%}), ``drink from can'' (\textcolor{OliveGreen}{+25\%}), ``drink from bottle'' (\textcolor{OliveGreen}{+22\%}), and ``walk'' (\textcolor{OliveGreen}{+15\%}). These improvements stem from the model's ability to capture fine-grained skeletal dynamics, enabling robust discrimination between actions with similar object interactions but distinct motion patterns. Notably, skeleton tokens demonstrate enhanced capability in temporal action modeling, particularly in sequential actions involving multiple body positions such as ``lay down'' (\textcolor{OliveGreen}{+18\%}), ``get up'' (\textcolor{OliveGreen}{+16\%}), and ``sit down'' (\textcolor{OliveGreen}{+17\%}). The skeleton cue also excels in capturing subtle motion differences in drinking actions (``drink from cup'' \textcolor{OliveGreen}{+20\%}, ``drink from can'' \textcolor{OliveGreen}{+25\%}, ``drink from bottle'' \textcolor{OliveGreen}{+22\%}), where the trajectory and orientation of movement are key discriminative features.

Our proposed MMPro framework, demonstrates significant synergistic effects, particularly in complex actions requiring both object and kinematic understanding. For instance, MMPro achieves significant improvements in ``use tablet'' (\textcolor{OliveGreen}{+58\%}), ``make coffee'' (\textcolor{OliveGreen}{+52\%}), ``breakfast take ham'' (\textcolor{OliveGreen}{+45\%}), and ``use laptop'' (\textcolor{OliveGreen}{+48\%}), where both object state changes and body motion patterns are crucial for accurate classification. The framework's effectiveness is particularly evident in ambiguous scenarios where individual modalities underperform, such as ``clean dishes'' (\textcolor{OliveGreen}{HOI: +18\%}, skeleton: \textcolor{OliveGreen}{+15\%}, MMPro: \textcolor{OliveGreen}{+25\%)}, ``cook stir'' (HOI: \textcolor{OliveGreen}{+22\%}, skeleton: \textcolor{OliveGreen}{+20\%}, MMPro: \textcolor{OliveGreen}{+32\%}), and ``pour from kettle'' (HOI: \textcolor{OliveGreen}{+20\%}, skeleton: \textcolor{OliveGreen}{+22\%}, MMPro: \textcolor{OliveGreen}{+30\%}).

In complex composite actions like ``make coffee'' (involving ``pour grains'' \textcolor{OliveGreen}{+28\%}, ``pour water'' \textcolor{OliveGreen}{+25\%)}, MMPro successfully captures both the object state transitions and the associated body movements, resulting in more accurate action classification. Our results demonstrate that the MMPro strategy successfully addresses the limitations of video only LLVMs, providing a more comprehensive framework for action understanding in complex real-world scenarios.

\section{Additional Qualitative Evaluation}
\label{sec:qual_eval}
\begin{figure*}[h]
    \centering
    \includegraphics[width=\textwidth, height=5.5cm]{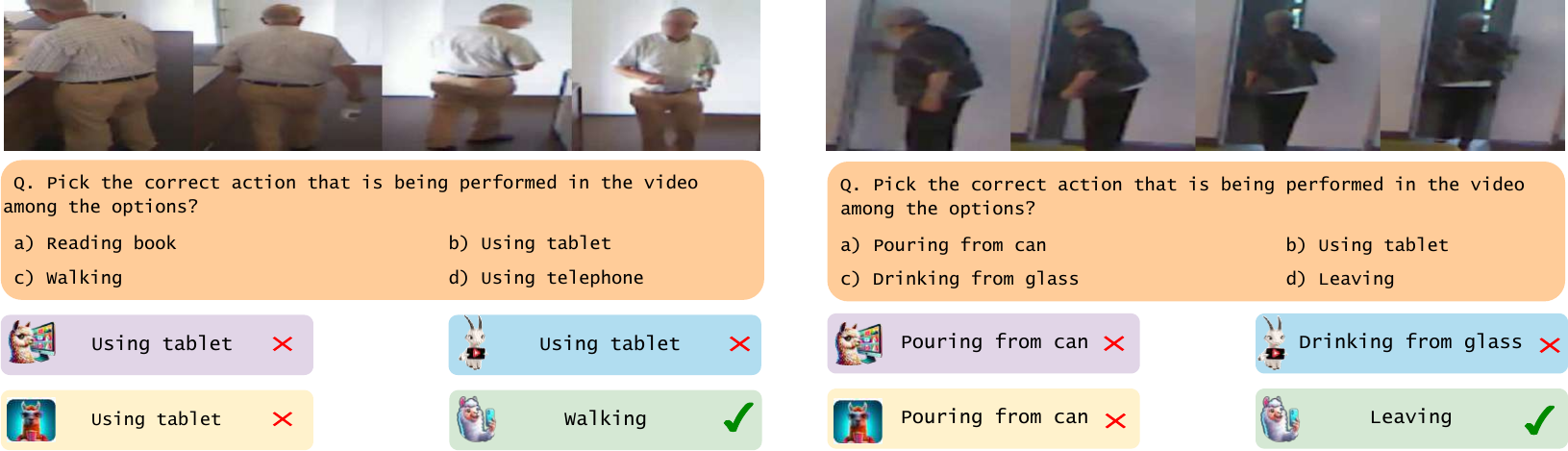} 
    \caption{Evaluation of ADL MCQ Action recognition task on Charades Dataset}
    \label{fig:Action_Recognition}
\end{figure*}

\begin{figure*}[h]
    \centering
    \includegraphics[width=\textwidth]{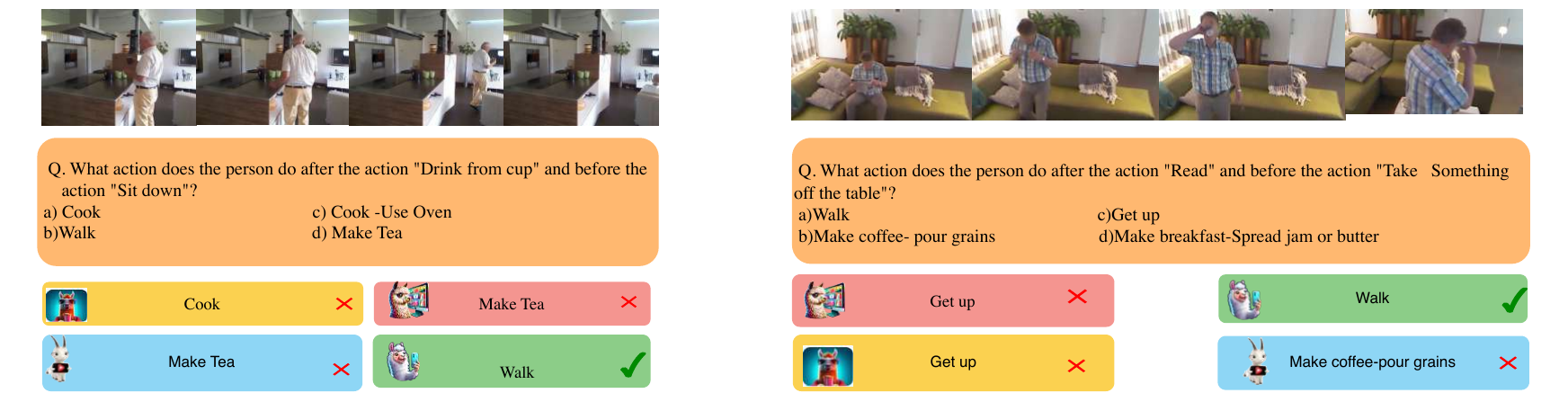} 
    \caption{Evaluation of ADL MCQ TC task on TSU dataset}
    \label{fig:Action_TC_TSU}
\end{figure*}

\begin{figure*}[h]
    \centering
    \includegraphics[width=\textwidth]{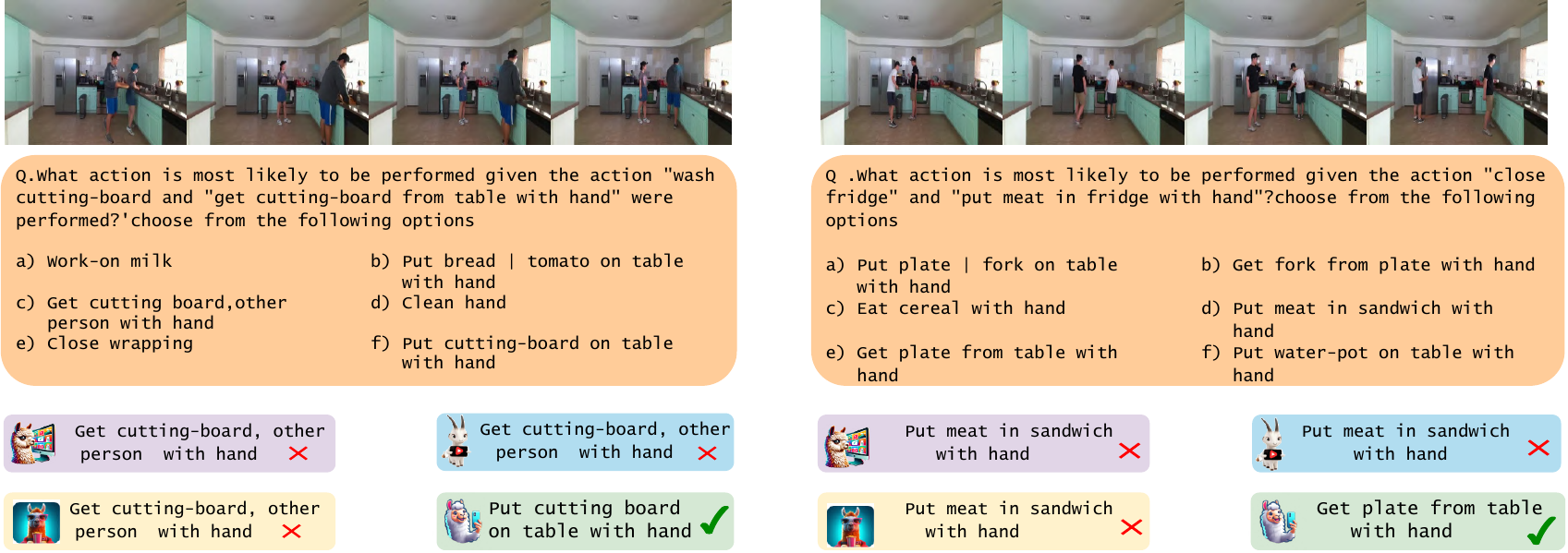} 
    \caption{Evaluation of ADL MCQ TC task on Lemma dataset}
    \label{fig:Action_TC_LEMMA}
\end{figure*}

\begin{figure*}[h]\vspace{-0.2in}
    \centering
    \scalebox{0.9}{
    \includegraphics[width=\textwidth]{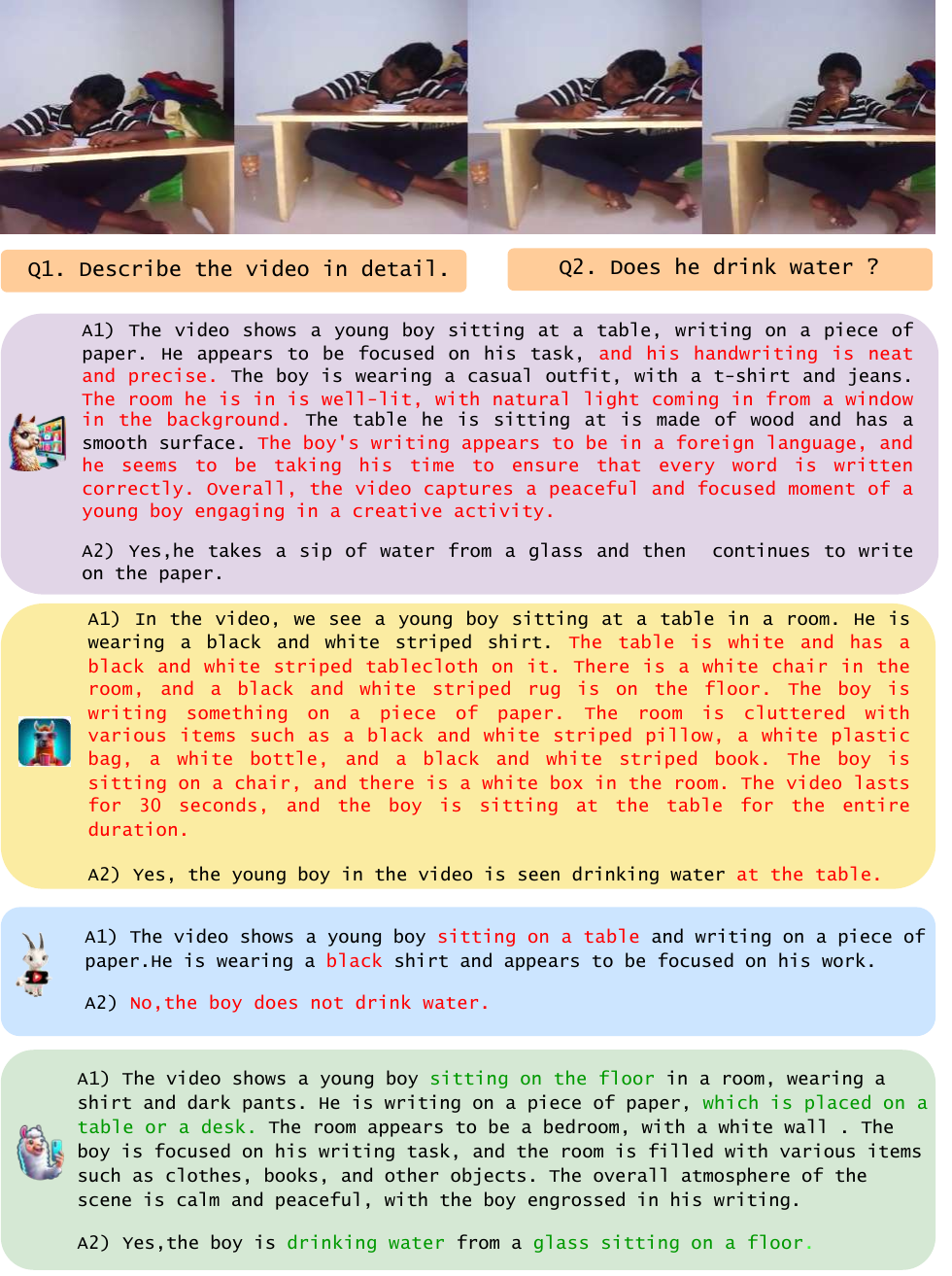} }
    \caption{Qualitative evaluation of video description on Charades Dataset. Text marked in \textcolor{red}{red} are incorrect. Text marked in \textcolor{green}{green} are correct.}
    \label{fig:Video_Description}
\end{figure*}

In this section, we provide qualitative evaluation of \modelname~and other state-of-the-art LLVMs for the tasks of ADL MCQ Action Recognition and ADL MCQ Temporal Completion, illustrated in Figures~\ref{fig:Action_Recognition}, \ref{fig:Action_TC_TSU}, and \ref{fig:Action_TC_LEMMA}. In Figure~\ref{fig:Video_Description}, we demonstrate the performance of \modelname~for Video Description Generation on the Charades dataset.

\begin{figure*}[h]
    \centering
    \includegraphics[width=\textwidth]{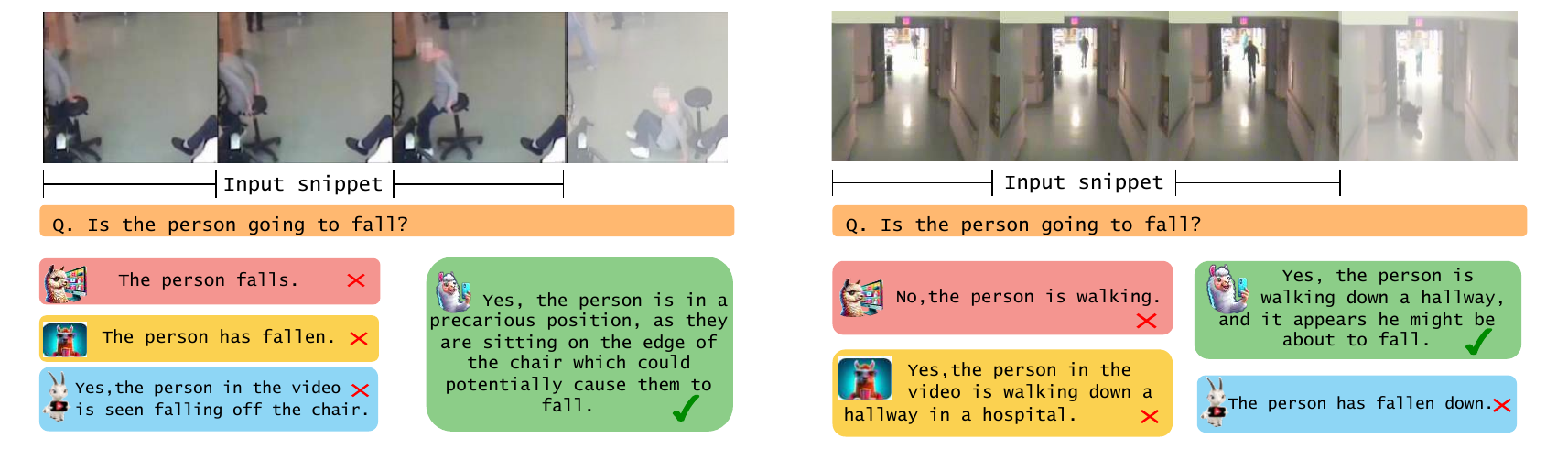} 
    \caption{The input snippet is the input video and the \textcolor{gray}{grey} part is omitted out, here the model needs to detect the \textcolor{gray}{greyed} action.}
    \label{fig:fall}
\end{figure*}

One of the applications of \modelname~is to monitor cognitive decline in geriatric patients through the action forecasting capabilities of our model. In this effort, we have qualitatively evaluated the model on videos of falls on long term care by the IMPL SFU~\cite{FallDetection}. The subjects in these videos are suffering from dementia, seizure, diabetes like diseases and the dataset contains 175 such falls. We slice the input video before the event of \textit{fall} and prompt \modelname~and other LLVM's to predict whether the person will fall or not. As illustrated in Figure~\ref{fig:fall}, our model outperforms the other LLVMs by predicting the fall correctly and by giving proper explanation of why the fall would occur highlighting its reasoning capabilities. While other models predict that the person \textit{"has fallen down"} and hallucinates the reasoning of the fall as well. 
\vspace{-0.1mm}


\vspace{-0.1mm}

\vspace{-0.1mm}

\vspace{-0.1mm}

\vspace{-0.1mm}

\section{LLM Prompts Used}
\label{sec:prompts}
In the following sections, we demonstrate the prompts used:

\subsection{Dense Captioning using GPT-3.5 Turbo}

\textcolor{navy}{\textbf{\{"role":"system"\}:}}
\texttt{"You will play two roles: a human asking questions related to describing a video and an intelligent chatbot designed for video description and dense captioning. Your task is to generate a detailed and descriptive paragraph based on the provided fragmented information about a video."}

\textcolor{navy}{\textbf{"\#\#TASK":}}
\texttt{"Users will provide fragmented descriptions of a video, and you will generate ONE conversation-like question and answer related to describing the video in detail.The question should ask to describe the video content in detail.The answer should be a paraphrased and well-structured paragraph based on the provided description, with a minimum of 150 words and a maximum of 300 words.When the provided information is short, aim for a 150-word description, and when the provided information is more detailed, aim for very long descriptions up to 300-word description."}

\textcolor{navy}{\textbf{"\#\#INSTRUCTIONS":}}
\texttt{"The question must be like a human conversation and focused on describing the video in detail.The answer must be a paraphrased version of the provided information, very detailed and descriptive, and within the specified word count.Combine the information from different sections of the video into a single coherent summary, ignoring any repetitions.Compare the information across all fragments of video and remove or ignore any inconsistent information and do not say the summary comes from different fragments of the video.Give more emphasis on the actions, the objects, and the colors of the background and the objects.Give the sequence of actions happening in the video and the objects the person interacts with."}

\textcolor{navy}{\textbf{\{"role":"user"\}:}}
\texttt{"The fragmented video description is: \{mega\_caption\}. Please generate the response in the form of a Python dictionary string with keys "Q" for question and "A" for answer. Each corresponding value should be the question and answer text respectively. For example, your response should look like this: \{"Q": "Your question here...", "A": "Your answer here..."\}. Emphasize that the answer should focus on describing the video content following the given instructions."}

\subsection{QA generation using GPT-3.5 Turbo: Prompt 1}

\textcolor{navy}{\textbf{\{"role":"system"\}:}}
\texttt{"You play two roles: a human asking questions related to summarizing a video and an intelligent chatbot designed for video summarization and dense captioning. Your task is video summarization. As an AI assistant, assume that you have watched the video and generated the provided caption as the summary of the video. Your task is to play the role of a human who asks three questions related to summarizing the video and then play the role of an AI assistant that provides paraphrased answers based on the video content and the provided caption."}

\textcolor{navy}{\textbf{"\#\#TASK":}}
\texttt{"Users will provide a caption of the video alongside dense caption describing detected objects in that scene, and you will generate a set of three conversation-like questions related to summarizing the video. The questions and answers can be very similar, but they should all focus on summarizing the video content. The answers should be paraphrased versions of the provided caption and the dense caption with the object detections. You have information about the video based on the provided caption and have summarized the events in it. You also have the dense caption with the object and scene details. Generate THREE different questions asking to summarize the video and provide detailed answers to each based on the caption and the dense caption."}

\textcolor{navy}{\textbf{"\#\#INSTRUCTIONS":}}
\texttt{"The questions must be like a human conversation and focused on summarizing the video. The answers must be paraphrased versions of the provided caption and the dense caption, and they should be detailed and descriptive." \\
"------" \\
"SAMPLE QUESTIONS:" \\
"- Can you provide a summary of the video?" \\
"- What are the main events in the video?" \\
"- Could you briefly describe the video content?"
}

\textcolor{navy}{\textbf{\{"role":"user"\}:}}
\texttt{"The video caption is: \{caption\}. The additional dense caption is: \{mega\_caption\}. Generate three different questions on summarizing the video, and provide answers that are paraphrased versions of the given caption and the dense caption. Please attempt to form question and answer pairs based on the two sets of text. Please generate the response in the form of a Python list of dictionary string with keys "Q" for question and "A" for answer. Each corresponding value should be the question and answer text respectively. For example, your response should look like this: [\{"Q": "Your first question here...", "A": "Your first answer here..."\}, \{"Q": "Your first question here...", "A": "Your first answer here..."\}, \{"Q": "Your first question here...", "A": "Your first answer here..."\}]. Emphasize that the questions and answers can be very similar, but they should all focus on summarizing the video content."}
\subsection{QA generation using GPT-3.5 Turbo: Prompt 2}

\textcolor{navy}{\textbf{\{"role":"system"\}:}}
\texttt{"You play two roles: a human asking questions related to a video and an intelligent chatbot designed for video summarization and dense captioning. Your task is extracting diverse video information. As an AI assistant, assume that you have watched the video and generated the provided caption as the summary of the video. Your task is to play the role of a human who asks three questions related to summarizing the video and then play the role of an AI assistant that provides paraphrased answers based on the video content and the provided caption."}

\textcolor{navy}{\textbf{"\#\#TASK":}}
\texttt{"Users will provide a caption of the video alongside dense caption describing detected objects,setting and details in that scene, and you will generate a set of three conversation-like questions related to the video. The questions and answers can be very similar, but they should all focus on the details of the video content. The answers should be paraphrased versions of the provided caption and the dense caption with the object and scene details. You have information about the video based on the provided caption and have summarized the actions in it. You also have the dense caption with the scene details. Generate THREE different questions asking the details of the video and provide detailed answers to each based on the caption and the dense caption and one question should be about what actions are happening which should come from captions of the video."}

\textcolor{navy}{\textbf{"\#\#INSTRUCTIONS":}}
\texttt{"The questions must be like a human conversation and focused on finding the intricate and unique details of the video. The answers must be paraphrased versions of the provided caption and the dense caption, and they should be detailed and descriptive. "
"------" \\
"SAMPLE QUESTIONS:" \\
"- What are the actions occuring sequentially in the video?" \\
"- What are the colors of the outfits of the person in the video?" \\
"- What are the objects in the scene?" \\
"- What is the person doing?"}

\textcolor{navy}{\textbf{\{"role":"user"\}:}}
\texttt{"The video caption is: \{caption\}. The additional dense caption is: \{mega\_caption\} Generate three different questions on the details of the video, and provide answers that are paraphrased versions of the given caption and the dense caption. Please attempt to form question and answer pairs based on the two sets of text. Please generate the response in the form of a Python list of dictionary string with keys "Q" for question and "A" for answer. Each corresponding value should be the question and answer text respectively. For example, your response should look like this: [\{"Q": "Your first question here...", "A": "Your first answer here..."\}, \{"Q": "Your first question here...", "A": "Your first answer here..."\}, \{"Q": "Your first question here...", "A": "Your first answer here..."\}]. Emphasize that the questions and answers can be very similar, but they should all focus on the various details of the video content and understanding what actions are happening. Include at least one question about the sequence of actions happening in the video."}

\subsection{skeleton Description Generation Prompt using GPT-3.5 Turbo}
\texttt{I have the coordinates that track the position of human joints throughout a video. I want to obtain the motion of each of these joints over time, using only these human joint coordinates. Here are the joint coordinates across observations: \{pose\_str\}. I want to know the general motion of these joints AND the amount of this motion (if the joint moved a lot, or only a small amount over the frames). Respond with a single sentence that INDEPENDENTLY describes the motion directions and amount for each joint over the entire video. Please start your reply for each joint with the name of the joint. What can you tell me about the motion and motion magnitudes of these joints? Describe the concrete direction of the motion of the joints, do not just say they move in many directions, but only describe how it moves and not its numerical coordinates. Do not forget to list the motion and amount of motion in two separate sentences. Begin each description with the name of the joint followed by a colon. Also include a sentence that captures the structure of the human body, such as the posture and position of the joints relative to one another
}

Here the pose\_str, is of the following format: \\
\texttt{In observation 0, the right knee is at (104, 201) and the left knee is at (106, 197) and the right hand is at (87, 162) and the left hand is at (134, 49) and the head is at (112, 40). In observation 1, the right knee is at (82, 208) and the left knee is at (87, 204) and the right hand is at (66, 167) and the left hand is at (122, 63) and the head is at (91, 38).....}

\subsection{Prompt to obtain Relevant Objects using GPT-3.5 Turbo}
\texttt{I have a video where the action "\{action\_label\}" is being performed by a human. I have detected all of the objects in the scene of this video, the objects I found are: \{found\_objects\}. I only want the objects that are relevant to the action "\{action\_label\}". From the list of detected objects, return only the objects that are relevant to the action being performed. It is crucial that the objects you return are contained in the list of objects I have given you, DO NOT create new objects or modify the names of the existing objects. Order the objects by their relevance to the action. IT IS OKAY TO NOT RETURN ANY OBJECTS IF NONE ARE RELEVANT, In this case respond with the string "None". The relevant objects are (return the objects separated by a comma) (never explain your decision).}


\section{Limitations}
\label{sec:limit}
While our approach works well with videos spanning a few seconds, it struggles with long videos. \modelname's preprocessing pipeline samples 100 frames per video. This sampling rate misses out key information in case of long videos, where there is a larger number of frames. To this end, for the task of generating Video Descriptions, we split the long videos in Toyota Smarthome Untrimmed into clips of 20 seconds each and generate descriptions for each clip. These clip-level descriptions are summarized using GPT3.5 Turbo to obtain a video-level description. However, this summarization step loses valuable information and hence fails to provide an accurate summary of the long video. Future work should explore an effective sampling strategy for long video understanding.

\section{Licensing and Intended Use}
\label{sec:license}
This paper introduces a large-scale dataset, \textbf{\datasetname}, comprising 100K untrimmed RGB video-instruction pairs, 3D skeletons, language descriptions, and action-conditioned object trajectories. The raw videos in \datasetname~ comprise content from NTURGB+D~\cite{NTU_RGB+D}, for which the original authors retain distribution rights for the clipped action videos. The scripts utilized to curate the dataset are open-sourced, facilitating the regeneration of the dataset. We will also provide comprehensive features, including image features extracted using CLIP, skeleton features derived from skeletonCLIP, and HOI features obtained through ObjectLM. We plan to release \datasetname~via an academic website for research, academic, and commercial use. The dataset is protected under the \textbf{CC-BY} license of Creative Commons, which allows users to distribute, remix, adapt, and build upon the material in any medium or format, as long as the creator is attributed. The license allows \datasetname~for commercial use. As the authors of this manuscript and collectors of this dataset, we reserve the right to distribute the data.
Additionally, we provide the code, data, and instructions needed to reproduce the main experimental baseline results, and the statistics pertinent to the dataset. We specify all the training details (e.g., data splits, hyperparameters, model-specific implementation details, compute resources used, etc.). Furthermore, we release the code and model weights of our proposed \textbf{L}arge \textbf{LA}nguage \textbf{VI}sion model for \textbf{D}aily \textbf{A}ctivities of \textbf{L}iving (\textbf{\modelname}), along with the features and instruction QA pairs for the combination videos. 
The \datasetname~dataset focuses on ADL and does not contain any personal data that can resemble evidence, reveal identification, or show offensive content.

The \datasetname~dataset can be used by multiple domain experts to advance research and development in various applications related to ADL. Its potential applications include, but are not limited to, assistive technologies, healthcare monitoring systems~\cite{ADLMONITORING}, smart homes~\cite{Chan2008}, robotics for assisted living, and instructional videos for ADL training and support. The dataset can also contribute to the development of AI-driven solutions that aim to improve the quality of life for individuals with disabilities, older adults, and those in need of daily assistance.
While we believe that the \datasetname~dataset has the potential to make a positive impact on society by enabling the development of technologies that support and enhance the lives of individuals, we acknowledge that, as with any technology, there is a possibility that the dataset or the ideas it presents could be misused or adapted for harmful purposes. However, as authors, we strongly oppose any detrimental usage of this dataset, regardless of whether it is by an individual or an organization, under profit or non-profit motivations. We pledge not to support any endeavors that could cause harm to individuals or society in relation to our data or the ideas presented herein.
Our intention is to foster research and innovation in the field of ADL analysis and support, ultimately contributing to the development of technologies that improve the quality of life for those who need assistance with daily activities. We encourage all users of the \datasetname~dataset to adhere to the highest ethical standards and to prioritize the well-being of individuals and society in their research and development efforts. 

\end{document}